\documentclass[10pt,journal,cspaper,compsoc]{IEEEtran}
\ifCLASSOPTIONcompsoc
\else
\fi
\ifCLASSINFOpdf
\else
\fi

\usepackage{times}
\usepackage{epsfig}
\usepackage{graphicx}
\usepackage{amsmath}
\usepackage{amssymb}
\usepackage{bm}
\usepackage[caption=false]{subfig}
\usepackage{hyphenat}
\usepackage{clrscode3e}
\usepackage{bm}
\usepackage{tikz}
\usetikzlibrary{fit,positioning}
\usepackage{capt-of}

\usepackage{xr}
\usepackage{flushend}

\newtheorem{theorem}{Theorem}[section]
\newtheorem{lemma}[theorem]{Lemma}

\newenvironment{proof}[1][Proof]{\begin{trivlist}
\item[\hskip \labelsep {\bfseries #1}]}{\end{trivlist}}

\newcommand{\qed}{\nobreak \ifvmode \relax \else
      \ifdim\lastskip<1.5em \hskip-\lastskip
      \hskip1.5em plus0em minus0.5em \fi \nobreak
      \vrule height0.5em width0.75em depth0.25em\fi}

\newcommand{\argmax}{\operatornamewithlimits{argmax}}
\newcommand{\argmin}{\operatornamewithlimits{argmin}}
\newcommand{\Loc}{l}
\newcommand{\Dsc}{d}
\newcommand{\Obj}{o}
\newcommand{\Mod}{m}
\newcommand{\Fg}{f}
\newcommand{\Hyp}{h}
\newcommand{\Abn}{a}
\newcommand{\Pix}{j}
\newcommand{\T}{t}
\newcommand{\Proj}{\text{Proj}}
\newcommand{\Aux}{\Phi}
\newcommand{\Lat}{z}
\newcommand{\Prot}{k}
\newcommand{\Shape}{w}
\newcommand{\Hpar}{b}
\newcommand{\Xpar}{c}
\newcommand{\NProt}{K}

\begin{document}
%
\title{Spatio-temporal Video Parsing for \\ Abnormality Detection}
%
%
%
%

\author{Borislav~Anti\'c~and~Bj\"orn Ommer
\IEEEcompsocitemizethanks{\IEEEcompsocthanksitem The authors are with
the Institute for Scientific Computing (IWR), Heidelberg University,
Germany.\protect\\
E-mail: borislav.antic@iwr.uni-heidelberg.de}
\thanks{}}

%
%

\markboth{}%
{Shell \MakeLowercase{\textit{et al.}}: Bare Demo of IEEEtran.cls for Computer Society Journals}
%


\IEEEcompsoctitleabstractindextext{%
\begin{abstract}

Abnormality detection in video poses particular challenges due to the infinite 
size of the class of all irregular objects and behaviors. Thus no (or by far 
not enough) abnormal training samples are available and we need to find 
abnormalities in test data without actually knowing what they are. Nevertheless, 
the prevailing concept of the field is to directly search for individual 
abnormal local patches or image regions independent of another. To address this 
problem, we propose a method for joint detection of abnormalities in videos by 
spatio-temporal video parsing. The goal of video parsing is to find a set of 
indispensable normal spatio-temporal object hypotheses that \emph{jointly} 
explain all the foreground of a video, while, at the same time, being supported 
by normal training samples. Consequently, we avoid a direct detection of 
abnormalities and discover them indirectly as those hypotheses which are needed 
for covering the foreground without finding an explanation for themselves by 
normal samples. Abnormalities are localized by MAP inference in a graphical 
model and we solve it efficiently by formulating it as a convex optimization 
problem. We experimentally evaluate our approach on several challenging 
benchmark sets, improving over the state-of-the-art on all standard benchmarks 
both in terms of abnormality classification and localization.

\end{abstract}

\begin{keywords}
Abnormality Detection, Video Analysis, Surveillance, Video Retrieval, Graphical 
Models, MAP Inference  
\end{keywords}}

\maketitle

\IEEEdisplaynotcompsoctitleabstractindextext

%
\IEEEpeerreviewmaketitle

\section{Introduction} \label{sec:introduction}

With the rapid growth of video data, there is an increasing need not only for 
recognition of objects and their behavior, but in particular for detecting the 
rare, interesting occurrences of unusual objects or suspicious behavior in the 
large body of ordinary data. Finding such abnormalities in videos is crucial for 
applications ranging from automatic quality control to visual surveillance. Due 
to the large within-class variability, recognizing normal objects is already 
difficult. Abnormality detection in crowded scenes, however, features the 
additional challenge that there exist infinitely many ways for an object to 
appear in unusual context (irregular object instance) or to behave abnormally 
(unusual activity). Most of these abnormal instances are beforehand unknown, as 
this would for instance require predicting all the ways somebody could cheat or 
break a law. It is therefore simply impossible to learn a model for all that is 
abnormal or irregular. Consequently, recent work on abnormality detection 
\cite{Mahadevan10} has focused on a setting where the training data contains 
only normal visual patterns. Thus a discriminative approach cannot be employed 
to directly localize irregularities in these benchmark datasets. But how can we 
find an abnormality \emph{without knowing what to look for}? In spite of this 
fundamental problem, the main paradigm in abnormality detection is at present to 
independently classify individual video patches \cite{Boiman07,Xiang05} or 
regions \cite{Zhong04}.

If we want to avoid the ill-posed problem of having to decide locally and 
separately about the abnormality of each image region, we need to abandon the 
standard approach of object detection, which aims at detecting all objects in a 
scene independently from one another. Since abnormality detection is typically 
concerned with videos from a static camera as in surveillance or industrial 
inspection, robust background subtraction algorithms \cite{Wright09} can be 
used for foreground/background segregation. Our goal is then to find a set of 
spatio-temporal object hypotheses that jointly explain all foreground pixels. 
This means that normal object hypotheses, which can be learned from the 
training data, are spread over the spatio-temporal volume of a video in
order to cover foreground pixels, while protruding into the background as
little as possible. These hypotheses need to explain the appearance and behavior 
of the underlying video regions. As objects are mutually overlapping in crowded 
scenes, the spatio-temporal placement of the object hypotheses can only be 
determined jointly. Thus, our aim is to simultaneously select those object 
hypotheses, which are necessary for explaining the foreground and to identify 
for each selected hypothesis the best matching instance from the set of all 
normal training samples. Abnormal objects are then those hypotheses which are 
required for explaining the foreground, but which themselves cannot be explained 
by a normal training sample. Video parsing \emph{jointly} infers all necessary 
object hypotheses, so that we can \emph{indirectly} discover all abnormal 
objects present in a scene without actually knowing what to look for.

Our video parsing approach consists of two stages. In the first phase, we detect 
a large number of object candidates in each video frame and then group them 
temporally into spatio-temporal object hypotheses. This shortlist of hypotheses 
is a superset of all candidates that might be eventually needed for parsing the 
video, i.e., it has a low false negative and high false positive rate. The 
object candidates in individual frames are obtained by running a discriminative
background classifier and keeping only those patterns which are very unlikely to 
be background. Subsequently, object candidates in individual frames are linked 
temporally according to their motion cues so as to establish the shortlist of 
spatio-temporal object hypotheses. In the second phase of video parsing, the 
goal is to select hypotheses from the shortlist that can explain the 
foreground, and to simultaneously find normal object instances that match those 
hypotheses. We formulate this as an inference problem in a graphical model 
whose goal is to maximize the probability of the foreground explanation in a 
video. The inference in the graphical model is cast as a convex optimization 
problem where the unknown variables indicate both, the selection of hypotheses 
from the shortlist and their corresponding normal object prototypes learned 
from the training videos. Correspondences between hypotheses and normal object 
prototypes are based upon their shape, location as well as their appearance and 
behavior. The probability of abnormality of each hypothesis necessary for 
explaining the foreground is then calculated using the results of inference. 
Beside identifying abnormal objects, video parsing also computes per-pixel 
probability of abnormality, which effectively segments abnormalities without 
having any training samples for them.

We evaluate our approach on novel benchmark datasets for abnormality detection
that feature highly crowded scenes. As an example, the UCSD \textit{ped1} and 
\textit{ped2} anomaly detection and localization datasets \cite{Mahadevan10} 
contain busy walkways teeming with walking pedestrians. Abnormalities are not 
staged, but they occur spontaneously and correspond to unusual objects (e.g., 
vehicles in a pedestrian zone) or behaviors (e.g., a person cycling across 
walkways) in the scene. The training data features only normal patterns with 
large intra-class variability, whereas the test set consists of normal and 
abnormal instances. Due to the small resolution of videos (a person in the 
scene is on average only $20$ pixels tall) and heavy occlusion between objects 
in the scene, learning models of visual patterns is difficult. We also increase 
the future utility of the UCSD \textit{ped1} dataset by completing the 
pixel-wise ground-truth annotation for all videos in the test set that 
previously existed only for a small subset. The experimental results show a 
significant performance gain of our spatio-temporal video parsing approach in 
comparison to other state-of-the-art methods for abnormality detection.

\section{Related Work} \label{sec:literature}

We discuss here the previous work on abnormality detection in videos. The 
related problem of object recognition and tracking in crowded 
scenes \cite{Brostow06,zhao08} aims at recognizing and tracking objects of a
\textit{known} class in a scene, whereas our goal is to detect abnormal 
objects, all of them being instances of an \textit{unknown} class. Therefore, 
object recognition and tracking are beyond the scope of this paper and the 
details on these topics can be found in \cite{Ommer09}. Majority of the work on 
abnormality detection relies on the extraction of semi-local features from 
video  
\cite{Lowe04,Laptev04,Jiang11,Roshtkhari13,Schuster11}, that are then used 
to train a normalcy model. Abnormalities are detected if the normalcy model does 
not fit the data. Some approaches \cite{Dee04,Zhong04} are based on manually 
specifying constraints that define the condition of normalcy, whereas other 
methods \cite{Xiang05,Xiang08,Basharat08,Wang07,Zhu13,Yang13} 
learn the normalcy model directly from data in unsupervised way. 

The approach of Adam et al. \cite{Adam08} focuses on individual activities 
occurring only in selected parts of a scene. Kim and Grauman \cite{Kim09} 
detect abnormalities using a spatio-temporal Markov random field that 
adapts to abnormal activities in videos. Loy et al. \cite{Loy10} 
use active learning methodology to integrate human feedback into the 
detection of abnormal events and behaviors. Unsupervised topic models are used 
for detection of abnormal behaviors in \cite{Wang09,Hospedales09}. Hospedales 
et al. \cite{Hospedales11} propose a semi-supervised multi-class topic model to 
classify and localize the subtle behavior in cluttered videos. Mahadevan et. al 
\cite{Mahadevan10} detect unusual objects in crowded scenes by jointly modeling 
the dynamics and appearance with mixtures of dynamic textures. Li et al. 
\cite{Li13} use the mixture of dynamic textures at multiple scales to 
detect abnormalities in a conditional random field framework. 

Kratz and Nishino \cite{Kratz09} develop a statistical model of local motion 
patterns in very crowded scenes to find abnormalities as local 
volumes with a large motion variation. Benezeth et al. \cite {Benezeth11} use 
low-level features to learn the co-occurrence matrix of normal behavior, and 
apply Markov random field to find deviating behaviors. Cong et al. \cite 
{Cong11} use sparse reconstruction cost implemented on a normal dictionary of 
local spatio-temporal patches to detect local and global abnormalities. 
Saligrama et al. \cite{Saligrama12} propose optimal decision rules for 
detecting local spatio-temporal abnormalities. An efficient sparse combination 
learning framework that achieves decent performance in the detection phase is 
proposed by Lu et al. \cite{Lu13}.

Instead of independently detecting abnormal regions in video as in other 
approaches, abnormalities are discovered indirectly after establishing a 
set of spatio-temporal hypotheses that provide complete explanation of the 
foreground. Previous approaches related to scene parsing differ in that a 
parametric scene \cite{Tu05,Ahuja08} or object model 
\cite{Kokkinos09,Fidler07,monroy12} or a non-parametric exemplar-based 
representation for objects \cite{Liu09,Malisiewicz09} can be constructed. In 
contrast to these methods we are not provided any training samples for the 
abnormalities we are searching for but we can leverage a foreground/background 
segregation. In contrast to our previous sequential video parsing 
\cite{AnticO11} that parsed video frames only spatially, one after another, the 
approach proposed in this paper performs a joint spatio-temporal parsing of 
video frames. This methodological extension is used to resolve both the spatial 
and temporal dependencies between objects in a scene. The new convex 
formulation of the inference process that improves upon the previous 
locally optimal inference method allows us to efficiently aggregate evidence 
from different frames and decide about their abnormalities in a globally 
optimal manner.

\begin{figure*}[t] 
\begin{center}
\subfloat[]{ \label{fig:all_hypotheses} 
\includegraphics[width=2in]{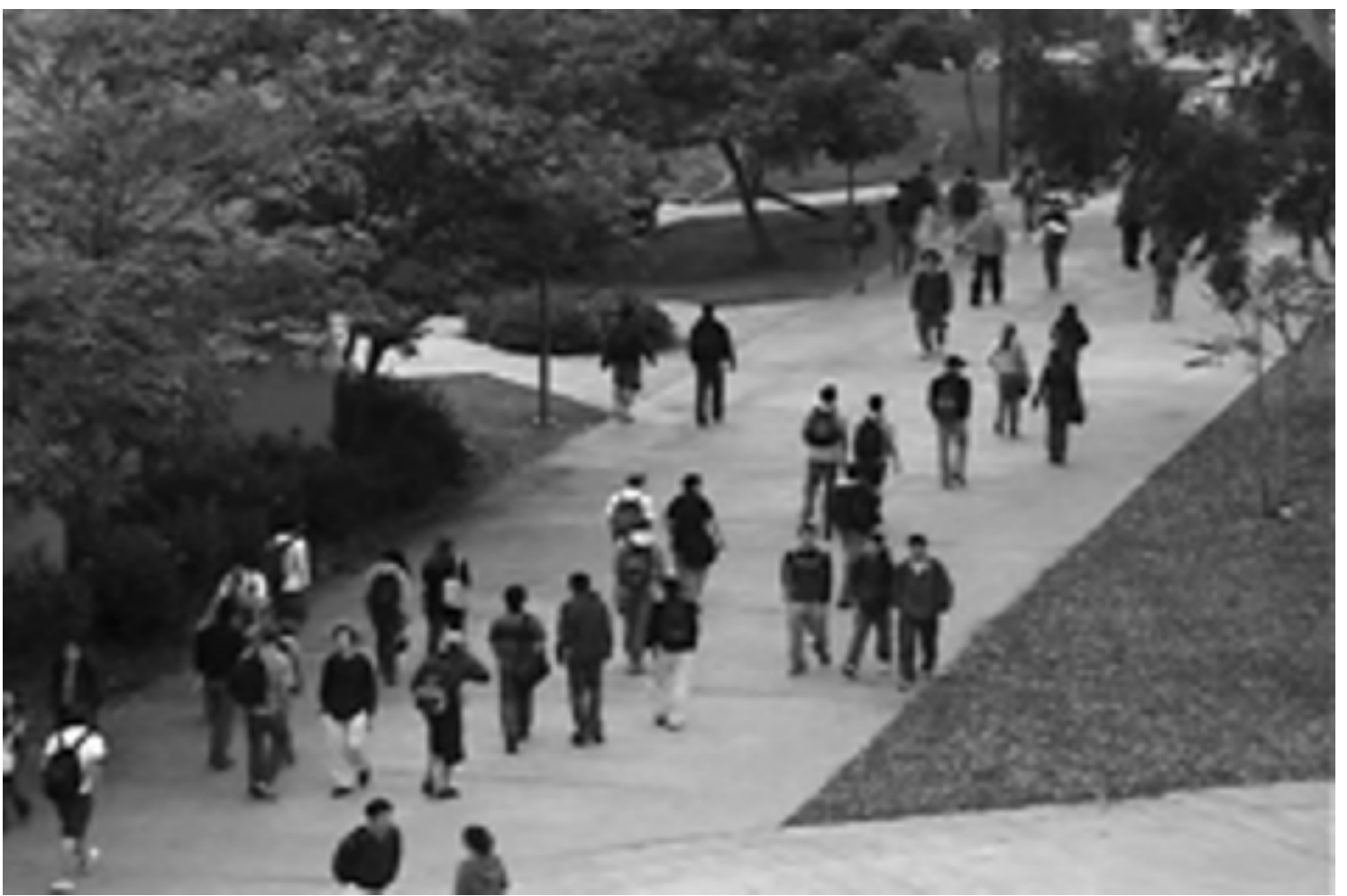}}
\hfill
\subfloat[]{ \label{fig:foreground_map} \setlength\fboxsep{0.5pt}
\fbox{\includegraphics[width=2in]{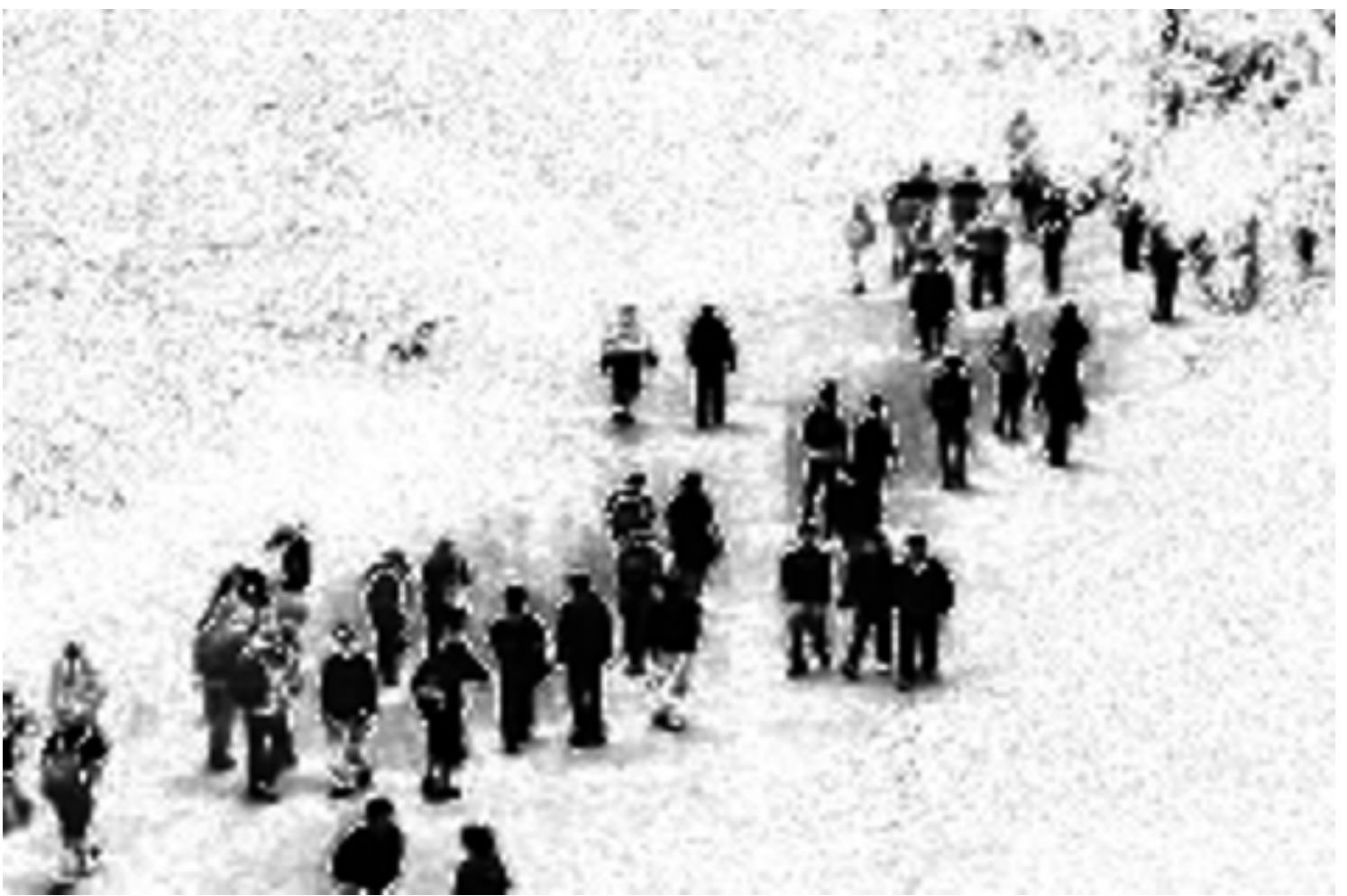}}}
\hfill
\subfloat[]{ \label{fig:all_hypotheses} 
\includegraphics[width=2in]{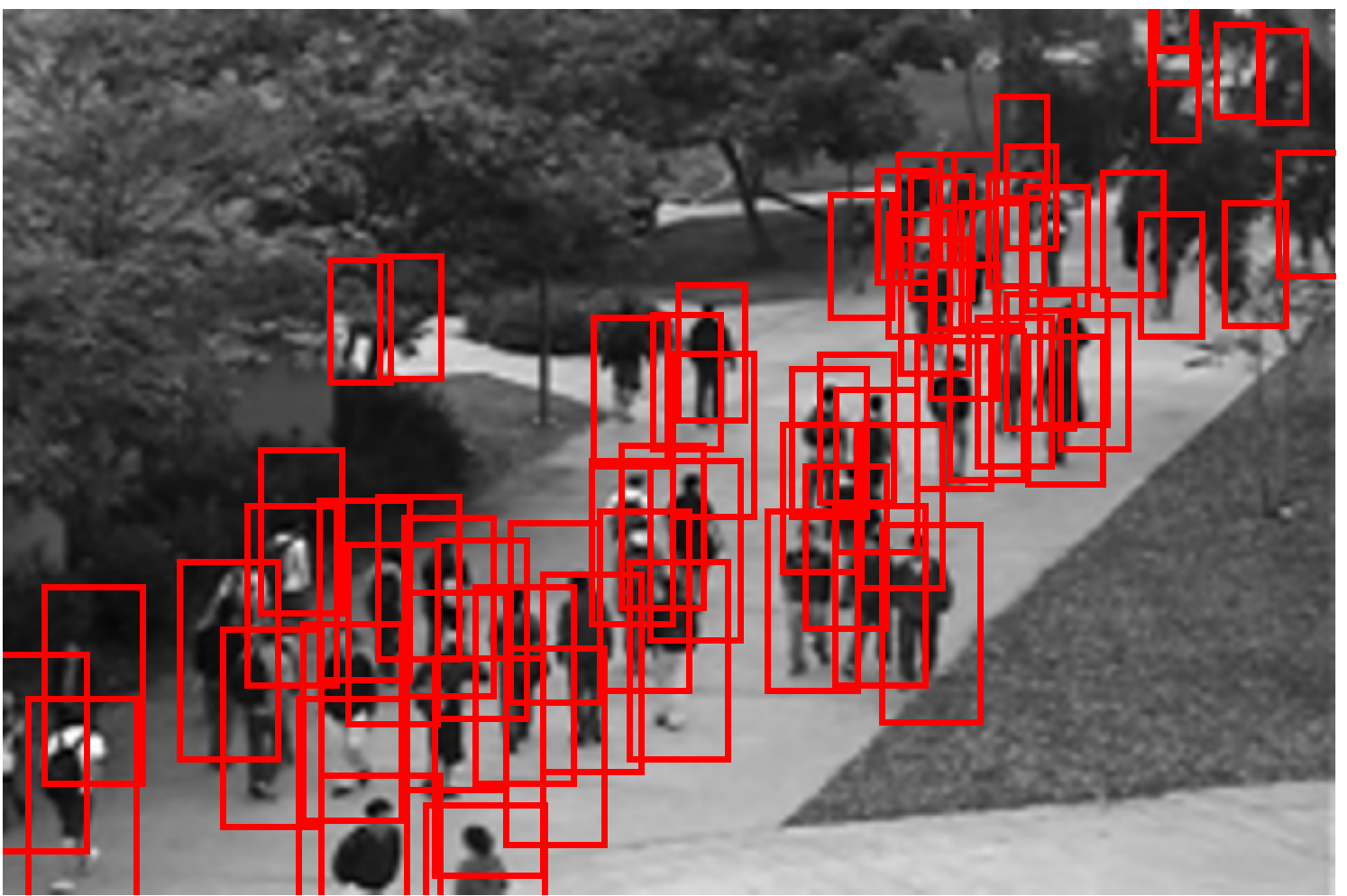}}
\hfill
\subfloat[]{ \label{fig:all_hypotheses} 
\includegraphics[width=2.3in]{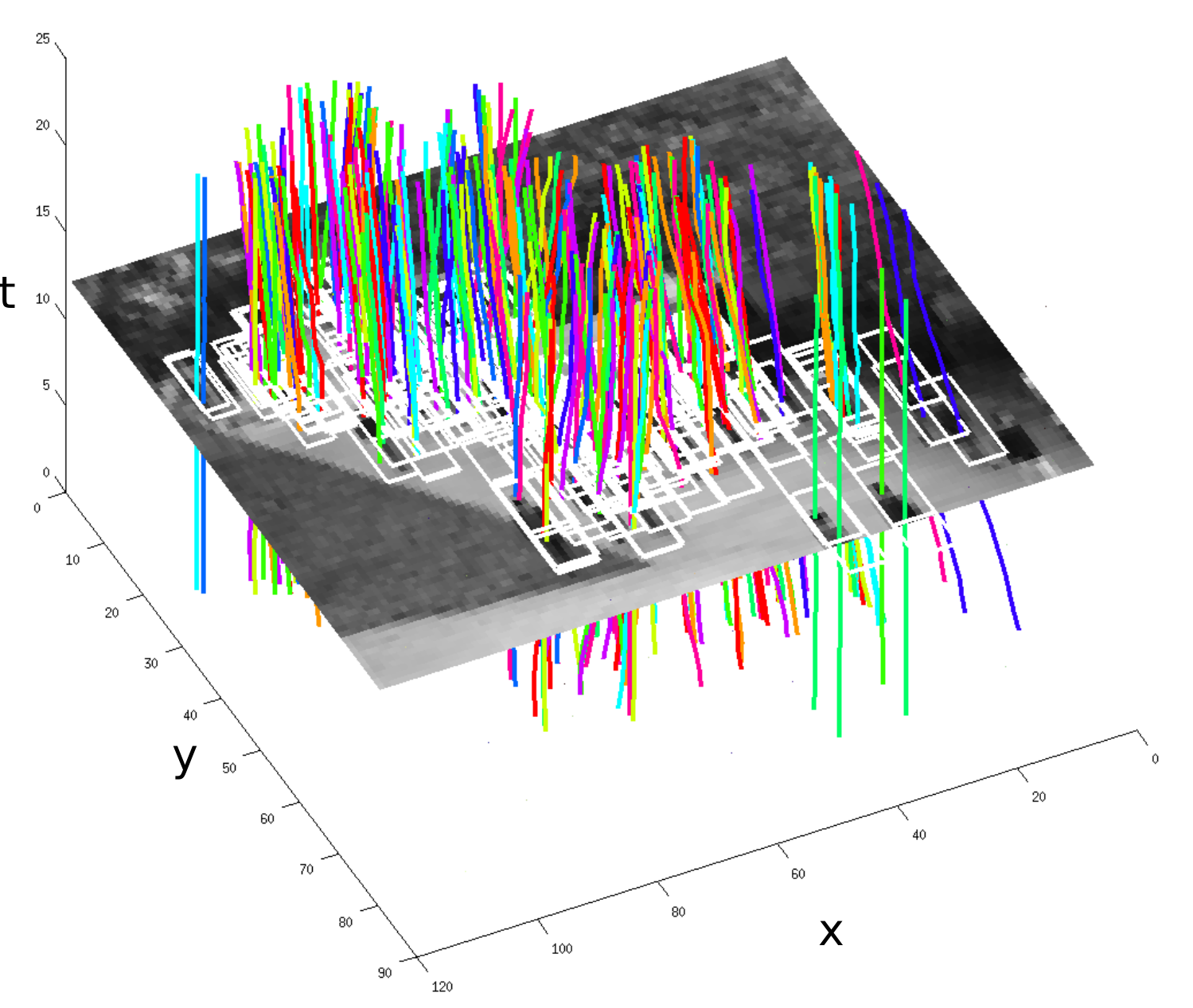}}
\subfloat[]{ \label{fig:parsed_hypotheses} 
\includegraphics[width=2.3in]{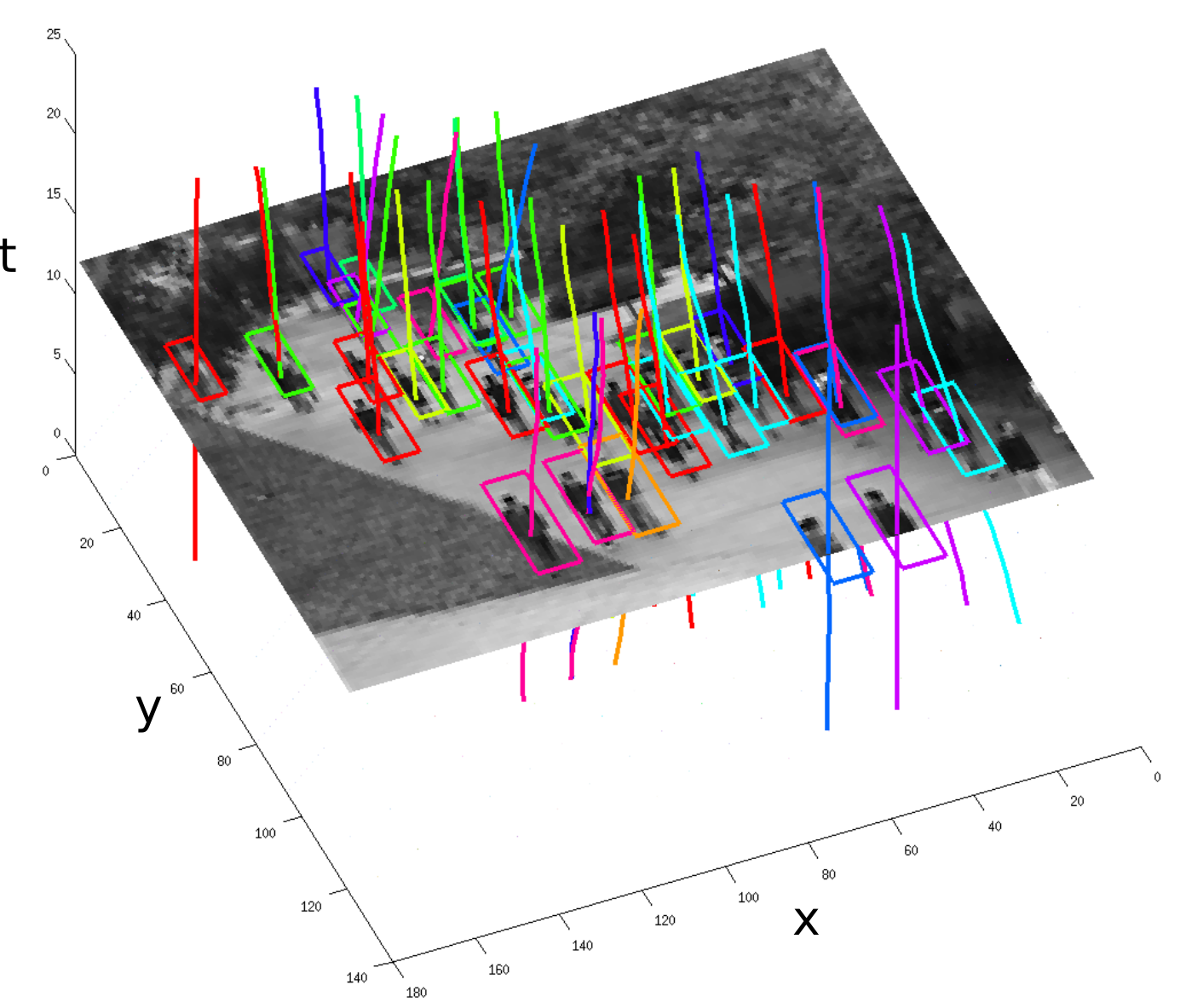}}
\hfill
\subfloat[]{ \label{fig:fitted_model} 
\raisebox{10mm}{\includegraphics[width=2in]{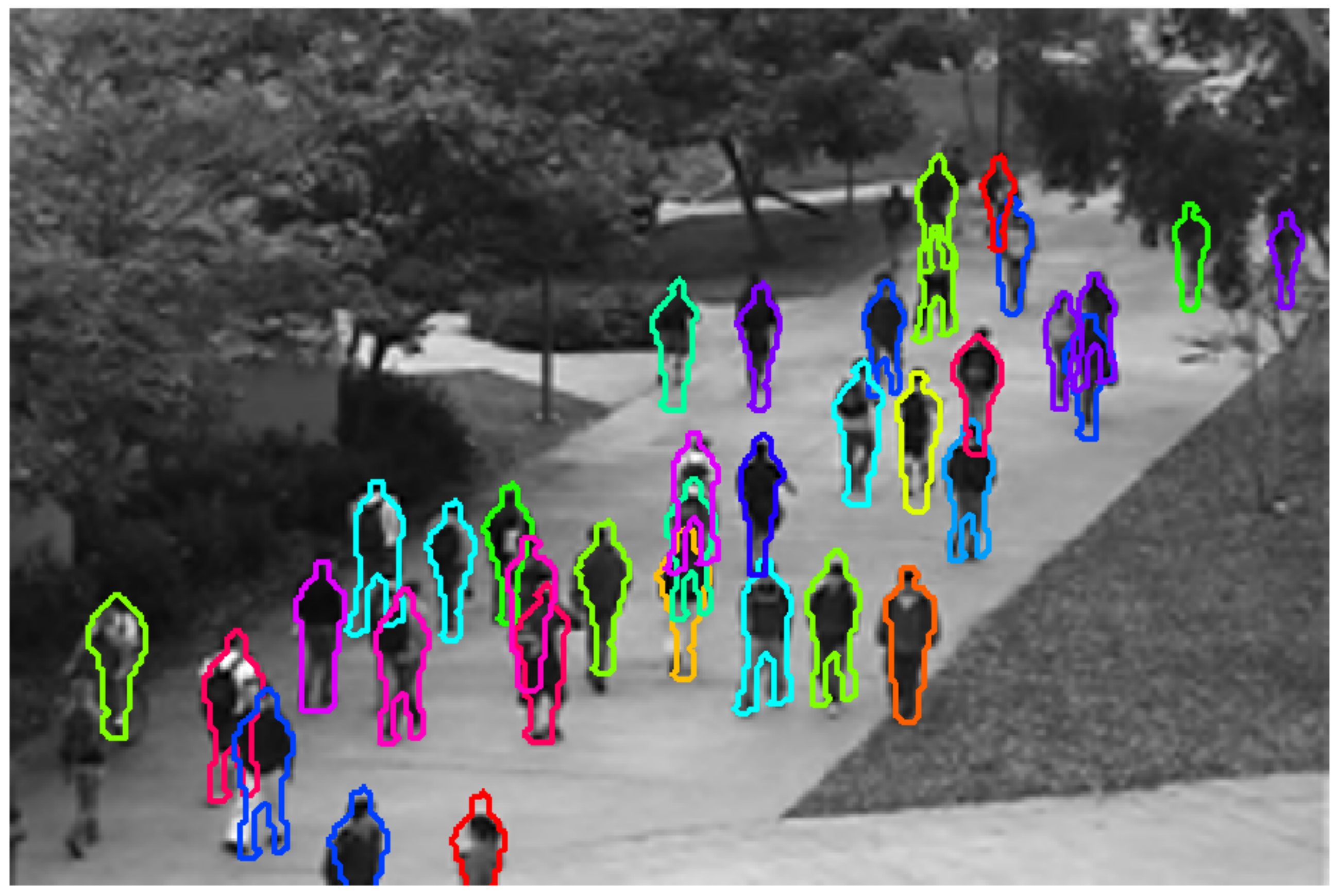}}}
\end{center}
\caption{Successive stages of the video parsing: (a) Source frame of a video. 
(b) Foreground probability map that needs to be explained by video parsing. (c) 
Object candidates found by inverted background detector. (d) Spatio-temporal 
object hypotheses found by temporal grouping serve as an input to the video 
parsing. (e) Subset of spatio-temporal object hypotheses that is selected by 
video parsing to explain the foreground pixels. (f) Normal object prototypes 
found by video parsing to explain the selected object hypotheses. Best 
viewed in color. \label{fig:hypotheses}} 
\end{figure*}

\section{Model for Spatio-temporal Video Parsing} \label{sec:parsing_model}

In case of a stationary camera, the foreground/background segregation becomes
feasible due to background subtraction. The foreground mask renders it then
possible to turn the abnormality detection problem into a task of video parsing.
The goal is thus to explain all the foreground of a video using object
hypotheses and to explain each hypothesis by an object model learned from
the set of normal training videos. The underlying statistical inference problem
has to be tackled jointly for all hypotheses, since hypotheses can explain each
other away. Abnormalities are then those hypotheses that are required to explain
the foreground but which themselves cannot be explained by any prototype from
the normal object model.

\textbf{Foreground segmentation}. Scenarios for abnormality detection often
involve the analysis of videos from static cameras. Background in such videos is
constant or changes slowly over time, hence it can be learned effectively from a
video. The resulting background model can then be applied to find all foreground
pixels in the video. The final foreground/background segmentation is represented
by a binary variable $\Fg _\Pix ^\T \in \{ 0 , 1 \}$ for all pixels $\Pix$ in 
frame $\T$. 

\textit{Background subtraction} assumes that each frame $I^\T$ of a video can be
expressed as the background model $B^\T$ plus a sparse vector $I^\T - B^\T$
whose nonzero elements are the foreground pixels. After stacking
successive video frames as columns in a matrix $I = \bigl[ I^{\T - \tau} \
\cdots \ I^\T \bigr]$, we want to find the low-rank background model $B$ such
that the sparsity inducing norm of the difference $I - B$ is the smallest
possible. Following the approach of Wright et al. \cite{Wright09}, we
approximate the rank of the matrix $B$ by a nuclear norm\footnote{Nuclear norm
is the sum of the singular values of the matrix and is a convex function.} $\|
\cdot \|_*$ and use $\ell_1$ as the sparsity inducing norm, so that the
background subtraction becomes the following convex optimization problem,
\begin{equation} \label{eq:bgsub}
 B = \argmin_{\tilde{B}} \| \tilde{B} \|_* + \| I - \tilde{B} \|_1.
\end{equation}

Now that we calculated the background model $B$, it can be used to find
all foreground pixels $\Pix$, $\Fg_\Pix^\T = 1$, as those that have a large
discrepancy between the observation $I_\Pix^\T$ and the background model
$B_\Pix^\T$. The probability that a pixel is foreground $P(\Fg_\Pix^\T=1)$
is obtained by the sigmoid transformation of the difference of pixel's intensity
and background model,
\begin{equation} \label{eq:bgprob}
 P(\Fg_\Pix^\T = 1) = 2 \Bigl(1 + \exp(- \lambda \| I_\Pix^\T - B_\Pix^\T
\|) \Bigr)^{-1} - 1.
\end{equation}
Pixels with foreground probability greater than $0.5$ are considered as
foreground, $\Fg_\Pix^\T = 1$, and others as background, $\Fg_\Pix^\T = 0$.

\textbf{Shortlist of Object Hypotheses}. For parsing the video, we need to
specify a list of spatio-temporal object hypotheses that is sufficient for
explaining foreground pixels in video. An input to our video parsing algorithm
consists of the most suitable object hypotheses for the task of foreground
explanation. In Sect. \ref{sec:shortlist} we explain the procedure for creating
a shortlist of object hypotheses that has a high recall, i.e. where the
majority of true-positive object hypotheses is included in the shortlist.
However, as the precision rate of the proposed shortlist is low, there will be
many superfluous hypotheses that are then explained away by others during video
parsing.

We assume that hypotheses from the shortlist span a time window $\{ \T - \tau,
\dots , \T \}$. Each hypothesis $\Hyp$ represents a spatio-temporal tube 
covering locations  $\Loc_\Hyp := (\Loc_\Hyp^{\T - \tau} \ \dots \ \Loc_\Hyp^\T 
)$. This is a trajectory of locations $\Loc_\Hyp^\T = ( x_\Hyp^\T \ y_\Hyp^\T \ 
s_\Hyp ^\T ) ^\top$, which specify the center $(x_\Hyp^\T, y_\Hyp^\T) $ and the 
scale $s_\Hyp^\T$ of a candidate object $\Hyp$ at time $\T$. The scale of an 
object represents its size relative to the size $(W, H)$ of the object model. 
The \textit{support region} of an object hypothesis $\Hyp$ at time $\T$ is the 
bounding box of size $(s_\Hyp^\T W, \ s_\Hyp^\T H)$, and the set of all pixels 
$\Pix$ that belong to it is denoted by $\mathcal{S}_\Hyp^\T$. 
 
The goal of video parsing is then to select a subset from the shortlist of
hypotheses that is both necessary and sufficient for explaining the foreground
of a test video while, at same time, finding normal object prototypes 
that explain the hypotheses of the subset (see Fig. \ref{fig:hypotheses}).

\begin{figure}[]
\centering
\begin{tikzpicture} 
  [->,>=latex,scale=1.0,auto=center,every node/.style={circle,draw}]
   \label{fig:graphmod}
  \node[fill=gray!20] (d) at (2,3)  {$\Dsc_\Hyp$};
  \node[fill=gray!20] (l) at (4,3)  {$\Loc_\Hyp$};
  \node (o) at (2,1.5)  {$\Obj_\Hyp$};
  \node (m) at (4,1.5)  {$\Mod_\Hyp$};
  \node (ah) at (3,0)  {$\Abn_\Hyp$};
  \node[fill=gray!20] (f) at (6,3) {$\Fg_\Pix^\T$};
  \node (ax) at (6,0)  {$\Abn_\Pix^\T$};
  \foreach \from/\to in {d/o,d/m,o/m,m/l,l/f,o/f,m/f,o/ah,m/ah,ah/ax,f/ax}
    \draw (\from) -- (\to);   
  \node[rectangle, inner sep=0mm, draw=none, fit= (o) (m) (d) (l)
(ah),label=below right:$\Hyp$, xshift=-3mm, yshift=-3mm] {};
  \node[rectangle, inner sep=4.4mm,draw=black!100, fit= (o) (m) (d) (l) (ah)]
{};
  \node[rectangle, inner sep=4.4mm,draw=black!100, fit= (f) (ax)] {};
  \node[rectangle, inner sep=0mm, draw=none, fit= (f) (ax),label=below
right:$\Pix$, xshift=-1mm, yshift=-14mm] {};
\end{tikzpicture}
\vspace{-0.4in}
\caption{Probabilistic graphical model of the spatio-temporal video parsing. 
The left plate contains all spatio-temporal hypotheses $\Hyp$ with their 
descriptors $\Dsc_\Hyp$ and locations $\Loc_\Hyp$. The right 
plate comprises all pixels $\Pix$ with their foreground labels $\Fg_\Pix^\T 
\in \{0,1\}$. By video parsing, we infer the set of hypotheses, $\Obj_\Hyp \in 
\{0,1\}$, that are necessary for explaining the foreground, and \textit{jointly} 
explain the selected hypotheses by the normal object prototypes $\Mod_\Hyp \in 
\{1, \dots, \NProt\}$. Finally, for each selected hypothesis $\Hyp$ we decide 
if it is abnormal, $\Abn_\Hyp \in \{0,1\}$, and also mark foreground 
pixels that belong to abnormal objects, $\Abn_\Pix^\T \in \{0,1\}$. 
\label{fig:graphmod}} 
\end{figure}
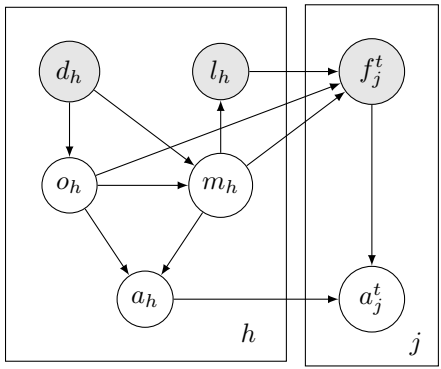

\textbf{Spatio-temporal object descriptor}. A spatio-temporal hypothesis $\Hyp$ 
matches its corresponding normal object prototype both in appearance and 
motion. Thus, we need a spatio-temporal descriptor $\Dsc_\Hyp$ 
to capture the essence of both appearance and motion of hypothesis $\Hyp$. We 
build a spatio-temporal descriptor $\Dsc_\Hyp := \bigl( \Dsc_\Hyp^{\T - \tau}  \ 
\dots \ \Dsc_\Hyp^\T \bigr) ^\top$ by concatenating frame-wise descriptors 
$\Dsc_\Hyp^\T$ calculated at each time $t$. Frame-wise object appearance is 
represented by the spatial derivatives of pixel's intensity in the support 
region $\mathcal{S}_\Hyp^\T$ of hypothesis $\Hyp$. Analogously, object motion is 
represented by the temporal derivatives of pixel's intensity. The appearance and 
motion representations are combined into frame-wise descriptor, 
\begin{equation} \label{eq:descriptor} 
 \Dsc_\Hyp^\T := \Bigl( \frac{ \partial I_\Pix^\T}{\partial x }, \
\frac{ \partial I_\Pix^\T} {\partial y}, \ \frac{ \partial I_\Pix^\T }{ \partial
t} \Bigr)_{\Pix \in \mathcal{S}_\Hyp^\T}.
\end{equation}
Since the spatio-temporal descriptor $\Dsc_\Hyp$ is long and redundant,  we 
build its compact representation by applying PCA transformation that projects 
onto eigen-space such that most of the signal variation is preserved (about 
$95\%$). 

\begin{figure}[tp]
  \centering
\subfloat{\includegraphics[width=0.45\textwidth,height=2in]
{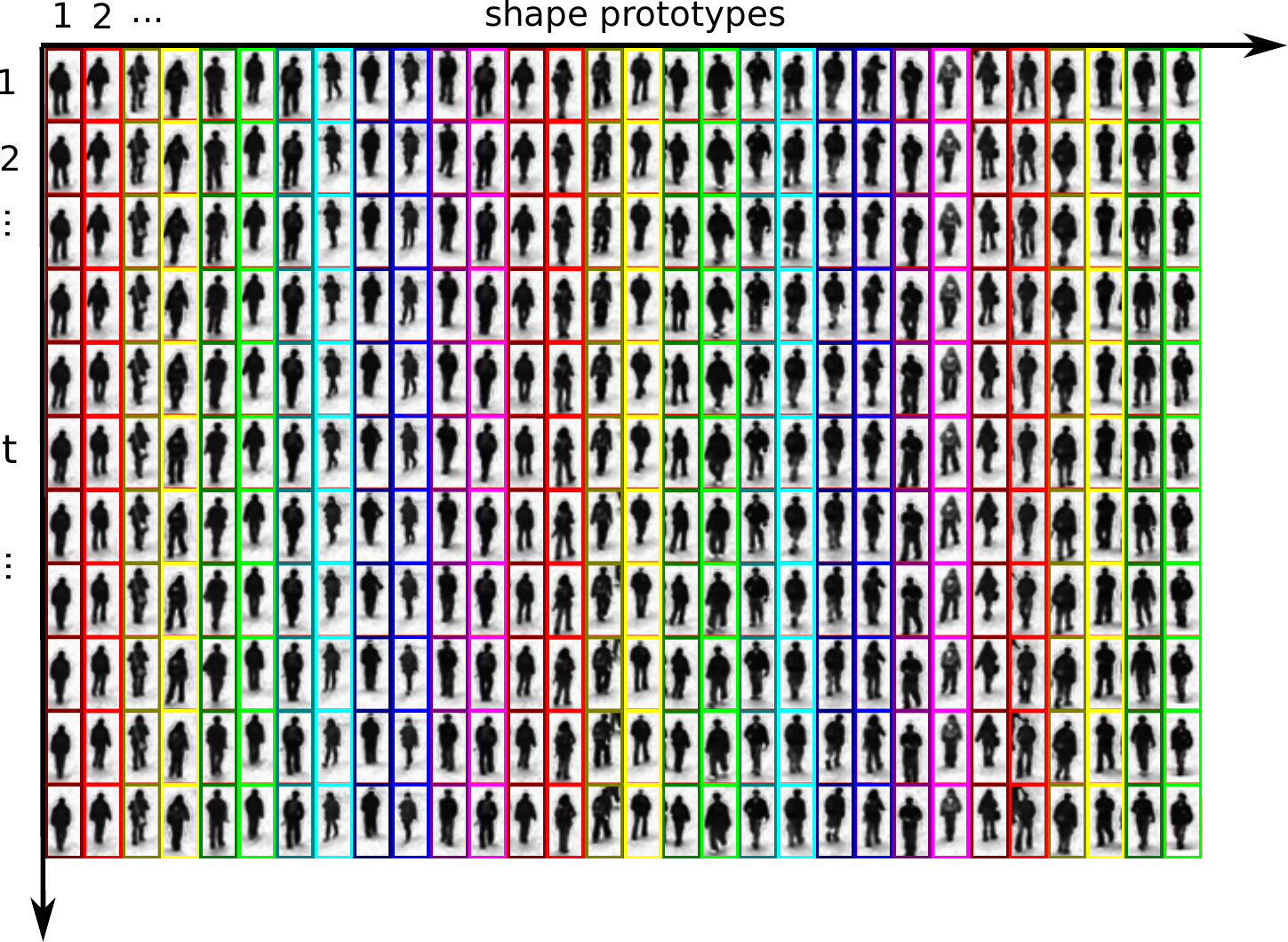}}
\caption{\label{fig:model_shapes} The normal object model consist of a set of 
spatio-temporal shape prototypes, each being a sequence that captures the 
temporal evolution of a particular shape. Prototypes are accompanied by the 
appearance and motion descriptors.}
\end{figure}

\textbf{Activating hypotheses needed for parsing}. Not all object hypotheses
from the shortlist are needed to explain foreground pixels in video. Video
parsing retains only the indispensable hypotheses that cannot be explained away 
by other hypotheses. Therefore, we use an indicator variable $\Obj_\Hyp \in 
\{0, 1\}$ for each hypothesis $\Hyp$ to designate the hypothesis as 
active/inactive. To initialize parsing, a discriminative classifier is 
trained to distinguish background spatio-temporal patterns from anything else. 
This background classifier computes the probability that hypothesis $\Hyp$ is 
background, $P(\Obj_\Hyp = 0 | \Dsc_\Hyp)$, which is then inverted to obtain the 
foreground probability. A hypothesis with high foreground probability can still 
become inactive if it gets explained away by others during video parsing.

\begin{figure}[t] 
\begin{center}
\subfloat[]{ \label{fig:firstfig} \label{fig:frameLabeling}
\includegraphics[width=0.4\textwidth]{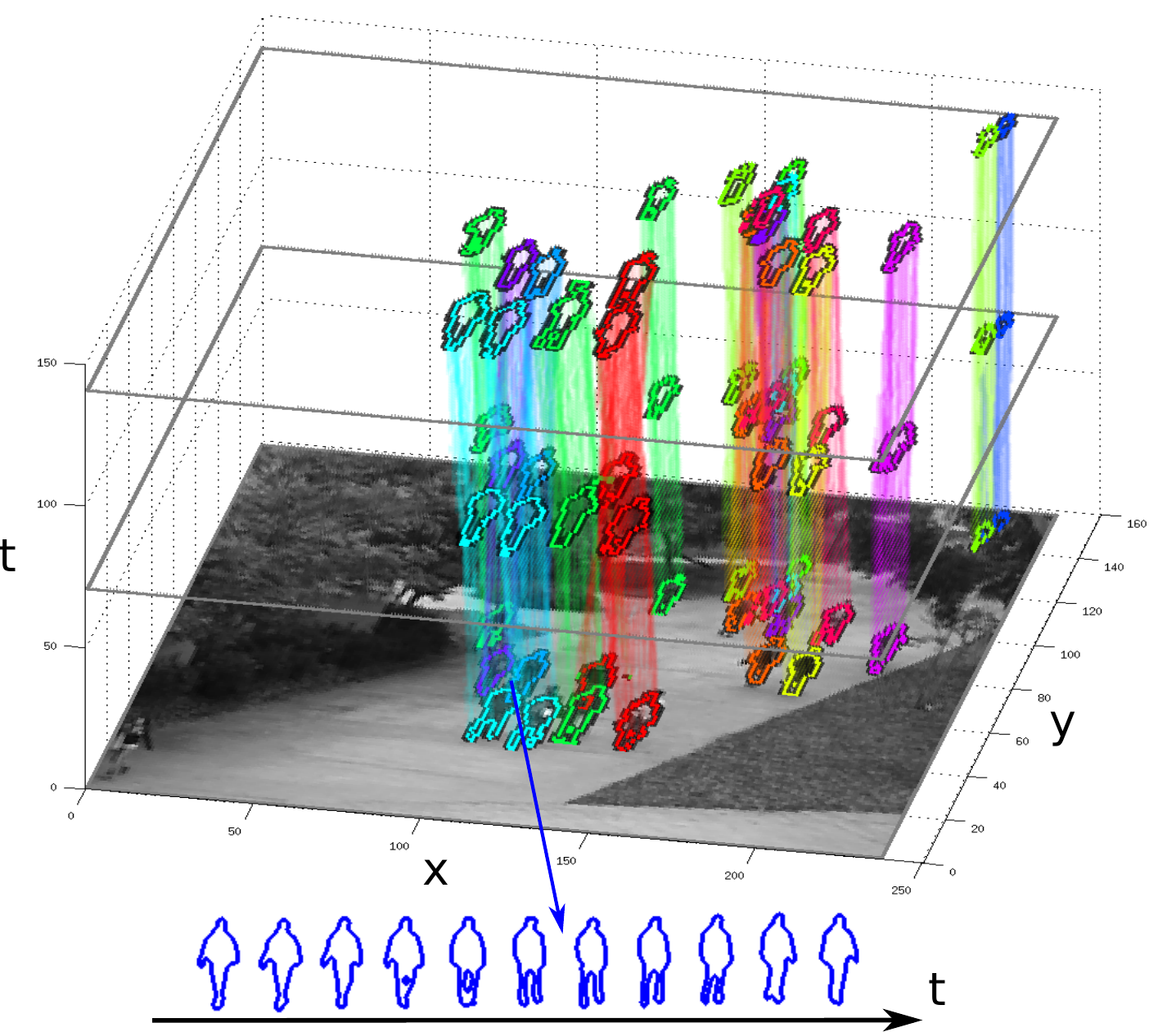}
}
\hfill
\subfloat[]{ \label{fig:firstfig} \label{fig:frameLabeling}
\includegraphics[width=0.4\textwidth]{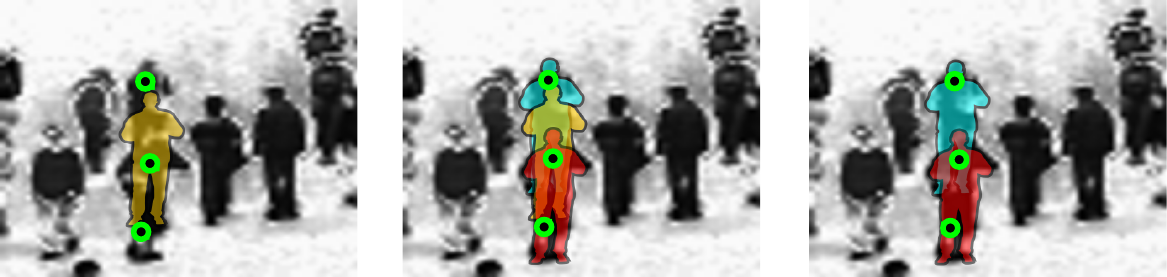}
}
\end{center}
\caption{ (a) Spatio-temporal tubes illustrate the hypotheses selected by video 
parsing. Normal shape contours that explain the hypotheses are shown 
overlaid. (b) Superfluous hypotheses are eliminated by the statistical 
inference of \textit{explaining away}. The idea is the following: Object 
hypothesis (yellow) is used at the beginning of video parsing to explain the 
foreground pixel in the middle. Other object hypotheses (red and blue) are 
introduced later to explain the top and bottom pixels. However, the pixel in 
the middle is also explained by new hypotheses, so that the original (yellow) 
hypothesis is not needed anymore and it can be eliminated. 
\label{fig:tubes}}
\end{figure}

\textbf{Matching with the object model}. Video parsing jointly explains
foreground pixels with object hypotheses, and active hypotheses $\{
\Hyp: \Obj_\Hyp = 1 \}$ with normal object prototypes learned from the 
training data. The object model consists of $\NProt$ normal object prototypes 
that represent a diversity of normal object's shape, appearance, and motion. 
Video parsing then determines for each selected hypothesis $\Hyp$ which of the 
$\NProt$ prototypes best explains it. The prototype that video parsing 
associates with hypothesis $\Hyp$ is indicated by the variable $\Mod_\Hyp \in 
\{1, \dots, \NProt\}$. Sect. \ref{sec:learning} explains in detail the learning 
of the normal object prototypes. For the time being, we assume that $\NProt$ 
normal object prototypes are provided as input to the parsing algorithm.

\begin{figure}[h] 
\begin{center}
\includegraphics[width=0.45\textwidth]{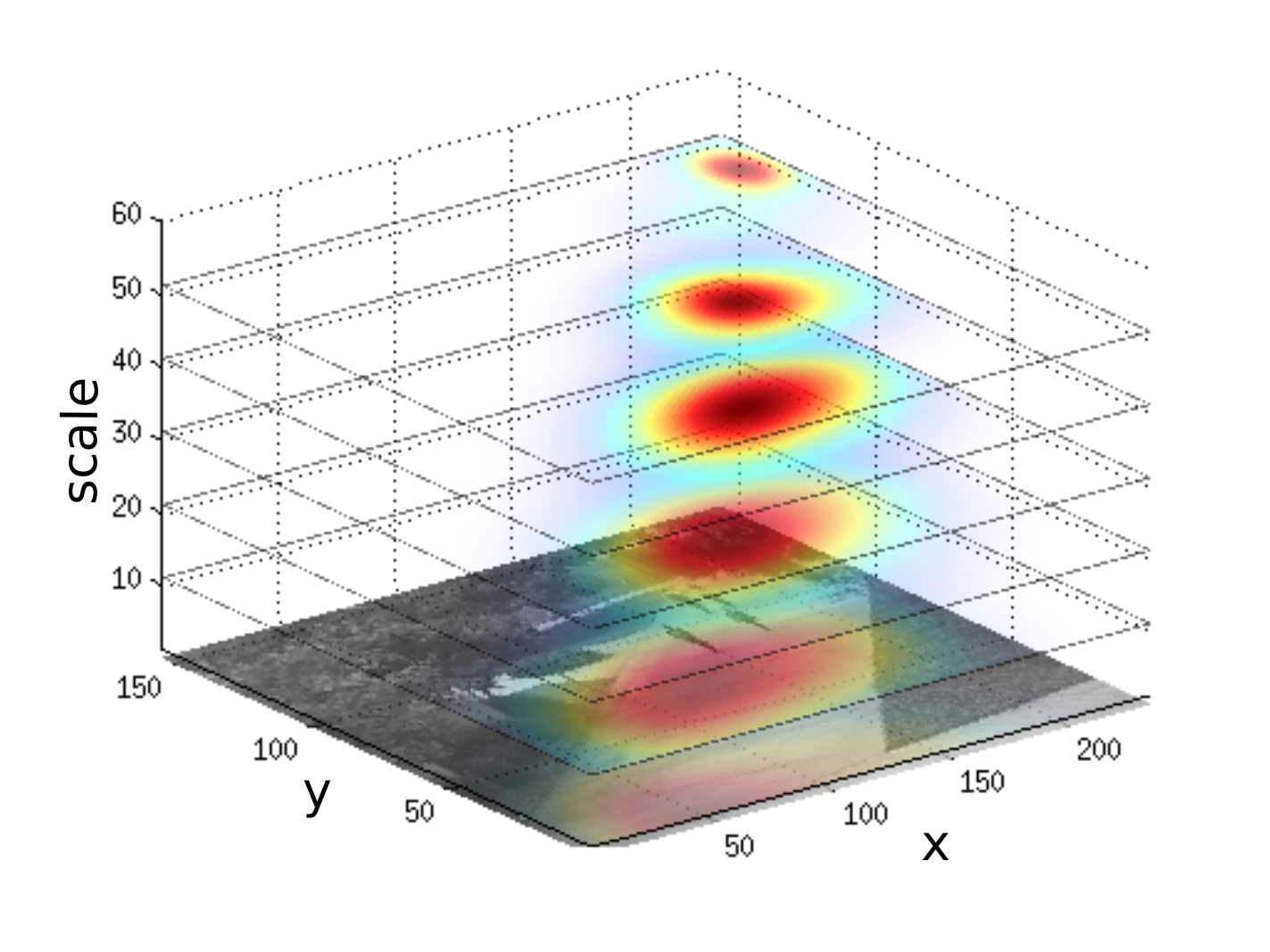}
\end{center}
\caption{\label{fig:parzen}The distribution of locations of normal object 
prototypes estimated by the Parzen windows at multiple scales (represented 
as horizontal slices).}
\label{fig:loc_distribution}
\end{figure}

For each hypothesis $\Hyp$ the best prototype $\Mod_\Hyp \in \{1, \dots, 
\NProt\}$ from the learned object model  is sought (Fig. 
\ref{fig:model_shapes}). For abnormal objects all prototypes will obviously 
have high matching costs. Consequently, the probability that prototype 
$\Mod_\Hyp$ is matched to a hypothesis $\Hyp$ in a query video depends on how 
similar they are in both appearance and motion, $\Delta( \Dsc_\Hyp, \Dsc_{ 
\Mod_\Hyp} )$. Here, $\Delta$ denotes a function that measures the 
distance of spatio-temporal descriptors in the corresponding feature 
space. Given the spatio-temporal descriptor $\Dsc_\Hyp$ of hypothesis $\Hyp$, 
the probability of matching prototype $\Mod_\Hyp$ with the hypothesis $\Hyp$ is 
the Gibbs distribution, 
\begin{equation} \label{eq:gibbs}
 P(\Mod_\Hyp | \Dsc_\Hyp) = \frac{1}{Z(\Dsc_\Hyp)} \exp \bigl(
- \beta \Delta (\Dsc_\Hyp, \Dsc_{\Mod_\Hyp}) \bigr),
\end{equation}
where $Z(\Dsc_\Hyp)$ is the partition function used to normalize the probability
distribution.

Moreover, normal objects typically occupy some location in a scene more
often than other, and also tend to move at a certain speed. For example,
cars are more likely to drive on roads than on sidewalks, whereas
pedestrians are more likely to walk on sidewalks. Consequently, the probability
of observing hypothesis $\Hyp$ that matches the prototype $\Mod_\Hyp$ depends on
its location $\Loc_\Hyp^\T$ and velocity $\Loc_\Hyp^\T - \Loc_\Hyp^{\T - 1}$,
\begin{equation} \label{eq:parzen} 
 P(\Loc_\Hyp | \Mod_\Hyp) \propto P_{\Mod_\Hyp}^{loc} (\Loc_\Hyp^\T
) \cdot P_{\Mod_\Hyp}^{vel} (\Loc_\Hyp^\T - \Loc_\Hyp^{\T - 1} ). 
\end{equation}
The normal location and velocity distributions $P_\bullet^{loc}$ and 
$P_\bullet^{vel}$ are learned for each of the $\NProt$ object prototypes using 
the Parzen window density estimator (see Fig. \ref{fig:parzen}).

Therefore, the probability that hypothesis $\Hyp$ matches to the normal object
prototype $\Mod_\Hyp$ is
\begin{alignat}{2}  \label{eq:matching}
 &P(\Mod_\Hyp | \Obj_\Hyp, \Dsc_\Hyp, \Loc_\Hyp) \propto
\Obj_\Hyp \cdot P(\Mod_\Hyp | \Dsc_\Hyp) \cdot P(\Loc_\Hyp | \Mod_\Hyp ).
\end{alignat}

\textbf{Explaining foreground pixels}. Video parsing selects hypotheses,
$\{ \Hyp: \Obj_\Hyp = 1 \}$, and finds corresponding normal object prototypes 
$\Mod_\Hyp$ to explain the foreground. The foreground probability of a pixel 
$\Pix$ depends on all hypotheses $\Hyp$ that overlap with pixel $\Pix$. Given 
the support regions $\mathcal{S}_\Hyp^\T$ of all hypotheses $\Hyp$,  $\{ \Hyp : 
\Pix \in \mathcal{S}_\Hyp\}$ is the set of hypotheses that cover the pixel 
$\Pix$. The probability that pixel $\Pix$ is background is equal to  the 
product of pixel's background probabilities for each single hypothesis $\Hyp$ 
that contains the pixel $\Pix$. Even if all hypotheses claim that pixel 
$\Pix$ is background, $P \bigl( \Fg_{\Pix}^\T = 1 | \Obj_\Hyp, \Mod_\Hyp, 
\Loc_\Hyp \bigr) = 0, \ \forall \Hyp$, we still allow it to be foreground with a 
small foreground probability $P_0 > 0$ . Thus, foreground probability of pixels 
$\Pix$ given all hypotheses is
\begin{alignat}{2} \nonumber
 P(\Fg_\Pix^\T &=1 | \{\Obj_\Hyp, \Mod_\Hyp, \Loc_\Hyp\}_\Hyp) = \\ 
 & 1 - (1 - P_0) \prod_\Hyp \Bigl( 1 - P(\Fg_\Pix^\T=1 | \Obj_\Hyp, \Mod_\Hyp,
\Loc_\Hyp) \Bigr).
\end{alignat}

The foreground probability given a single hypothesis $\Hyp$, $P( \Fg_\Pix^\T=1 |
\Obj_\Hyp, \Mod_\Hyp, \Loc_\Hyp ) $, depends on the shape of the 
corresponding normal object prototype $\Mod_\Hyp$. In the training data, the 
prototype $\Mod_\Hyp$ covers pixels $\Pix^\prime$ with some probability 
$P_{\Mod_\Hyp} (\Fg_{\Pix^\prime}^\T = 1)$. Thus, the foreground probability of 
pixel $\Pix$ under hypothesis $\Hyp$ is obtained by taking its corresponding 
object prototype $\Mod_\Hyp$ and ``pasting'' the foreground probability of 
$\Mod_\Hyp$ at the location of $\Hyp$. The model now needs to be brought into 
the reference frame of $\Hyp$ by scaling and translating it, i.e. $\Loc_\Pix^\T 
= s_\Hyp^\T \cdot \Loc_{\Pix^\prime}^\T + (x_\Hyp^\T \ y_\Hyp^\T )^\top$. Then 
the foreground probability of pixel $\Pix$ given $\Hyp$ becomes 
\begin{alignat}{2} \nonumber
 &P(\Fg_\Pix^\T=1 | \Obj_\Hyp, \Mod_\Hyp, \Loc_\Hyp) = \Obj_\Hyp \cdot
\mathbf{1} [ \Pix \in \mathcal{S}_\Hyp^\T ]  \\ \label{eq:pasting}
&\cdot \sum_{\Pix^\prime}  \mathbf{1} [ \Loc_\Pix^\T = s_\Hyp^\T \cdot
\Loc_{\Pix^\prime}^\T + (x_\Hyp^\T \ y_\Hyp^\T )^\top ] \cdot
P_{\Mod_\Hyp} (\Fg_{\Pix^\prime}^\T = 1).
\end{alignat}
Here $\mathbf{1}[\cdot]$ denotes the indicator function. In Eq.
\ref{eq:pasting} the foreground probability of pixel $\Pix$ is set to zero
if hypothesis $\Hyp$ is inactive, $\Obj_\Hyp = 0$, or the pixel $\Pix$ does not
belong to the support region of hypothesis $\Hyp$, $\Pix \notin \mathcal{S}
_\Hyp^\T$.

\section{Inference by Foreground Parsing}

The goal is now to estimate which of the hypotheses are actually needed for
explaining the foreground and to find a matching normal object prototype for 
each hypothesis. For abnormal hypotheses Eq. \ref{eq:matching} will yield low
probabilities. If foreground $\Fg_\Pix^\T=1$ is observed and the pixel is
covered by a hypothesis $\Hyp$, and no other hypothesis can be found that 
could explain the presence of the foreground at that pixel, then the 
probability of activation of hypothesis $\Hyp$ increases. This leads to the 
statistical inference of \textit{explaining away}. For an observed variable 
$\Fg_\Pix^\T$ different hypotheses $\Hyp$ that share the same pixel $\Pix$ 
become statistically dependent so that the absence of one hypothesis can 
dictate the presence of another (see Fig. \ref{fig:tubes}). 

\subsection{Joint Inference by MAP}

Based on the foreground segmentation mask $\Fg_\Pix^\T$ and the shortlist of
hypotheses $\Hyp$ with spatio-temporal descriptors $\Dsc_\Hyp$ and trajectories
$\Loc_\Hyp$, we need to jointly infer all hidden variables $\{\Obj_\Hyp,
\Mod_\Hyp\}_\Hyp$ in our graphical model (Fig. \ref{fig:graphmod}). Following a 
maximum a posteriori (MAP) approach yields a set of hypotheses that best 
explain the foreground and are themselves explained by the normal object 
prototypes,
\begin{alignat}{3} \nonumber \label{eq:map}
 &\{\bar{\Obj}_\Hyp, \bar{\Mod}_\Hyp\}_\Hyp = \max_{\{\Obj_\Hyp, 
\Mod_\Hyp\}_\Hyp} P(\{\Obj_\Hyp, \Mod_\Hyp\}_\Hyp | \{\Dsc_\Hyp, 
\Loc_\Hyp\}_\Hyp, \{\Fg_\Pix^\T\}_\Pix) \\ 
 &\propto \prod_\Pix P \bigl( \Fg_\Pix^\T | \{ \Obj_\Hyp, \Mod_\Hyp,
\Loc_\Hyp\}_\Hyp \bigr) \prod_\Hyp  P(\Obj_\Hyp | \Dsc_\Hyp) P(\Mod_\Hyp |
\Obj_\Hyp, \Dsc_\Hyp, \Loc_\Hyp). 
\end{alignat}

Instead of explicitly maximizing the posterior probability, we take a
negative logarithm of Eq. \ref{eq:map} and thereby obtain the energy function
$J(\cdot)$ which is then minimized. Furthermore, we decompose the energy
function $J(\cdot)$ into two terms, $J_\Pix(\cdot)$ covering the explanation of
foreground pixels $\Pix$, and $J_\Hyp(\cdot)$, which involves the explanation of
hypotheses $\Hyp$ by the normal object prototypes,
\begin{alignat}{2} \nonumber 
 &J(\{\Obj_\Hyp, \Mod_\Hyp\}_\Hyp) := \underbrace{ - \sum_\Pix  \log
P(\Fg_\Pix^\T | \{ \Obj_\Hyp, \Mod_\Hyp, \Loc_\Hyp\}_\Hyp )}_ {=:
J_\Pix(\{\Obj_\Hyp,\Mod_\Hyp\}_\Hyp)}  \\ \label{eq:J}
 &  \underbrace{ - \sum_\Hyp \Bigl( \log P(\Obj_\Hyp| \Dsc_\Hyp) + \log
P(\Mod_\Hyp | \Obj_\Hyp, \Dsc_\Hyp, \Loc_\Hyp) \Bigr) }_{=: 
J_\Hyp(\{\Obj_\Hyp,\Mod_\Hyp\}_\Hyp )}.
\end{alignat}

To find the MAP solution, we introduce a parsing indicator $\Lat_{\Hyp, \Prot} 
\in \{0, 1\}$, that equals one if hypothesis $\Hyp$ is active, $\Obj_\Hyp 
= 1$, and their corresponding normal object prototype is $\Mod_\Hyp = \Prot$,
\begin{equation} \label{eq:subst}
 \Lat_{\Hyp, \Prot} := \Obj_\Hyp \cdot \mathbf{1} [ \Mod_\Hyp = \Prot ], \
\forall \Hyp, \ \forall \Prot \in \{1, \dots, \NProt \}.
\end{equation}
To keep the notation simple, let the vector $ \mathbf{\Lat}_\Hyp := ( 
\Lat_{\Hyp,1} , \dots , \Lat_{\Hyp, \NProt} ) ^ \top $ denote the 
parsing indicators of hypothesis $\Hyp$, and the vector $\mathbf{\Lat} := \{ 
\mathbf{\Lat}_\Hyp \} _\Hyp$ denote the parsing indicators of all hypotheses 
together. The following lemma now states that the hypotheses explanation 
$J_\Hyp(\cdot)$ can be expressed as a linear function of the 
parsing indicator $\mathbf{\Lat}$.

\begin{lemma} \label{thm:hyp_explanation}
The hypotheses explanation term $J_\Hyp (\{\Obj_\Hyp, \Mod_\Hyp\}_\Hyp)$ in 
Eq. \ref{eq:J} is a linear function of the parsing indicator $\mathbf{\Lat}$, 
i.e.
\begin{equation} \label{eq:J_h}
 J_\Hyp (\{\Obj_\Hyp, \Mod_\Hyp\}_\Hyp) = \mathbf{\Hpar}^\top \mathbf{\Lat} + 
\Hpar_0,
\end{equation}
where the parameter vector $\mathbf{\Hpar} = 
\{ \Hpar_{\Hyp,\Prot}\}_{\Hyp,\Prot}$ and scalar $\Hpar_0$ do not depend on the 
parsing indicator $\mathbf{\Lat}$. The proof of Lemma \ref{thm:hyp_explanation} 
is given in Appendix \ref{app:derivation_Jh}.
\end{lemma}

To express the foreground explanation term $J_\Pix(\cdot)$ as a function of the 
parsing indicator $\mathbf{\Lat}$, we first define a function $\Aux_{\Fg_ 
\Pix^\T}(\cdot)$ that is parametrized by the foreground value $\Fg_\Pix^\T$ of 
pixel $\Pix$,
\begin{equation} \label{eq:helper}
 \Aux_{\Fg_\Pix^\T} (x) := (1-\Fg_\Pix^\T)  x - \Fg_\Pix^\T \cdot \log \bigl( 1-
e ^ {-x} \bigr), \ x > 0.
\end{equation}
The introduced function $\Aux_{\Fg_\Pix^\T}(\cdot)$ is convex as we show in the 
following lemma.

\begin{lemma} \label{thm:convexity}
The function $\Aux_{\Fg_\Pix^\T}(x), \ x > 0$ (Eq. \ref{eq:helper}) is convex 
for nonnegative values of the parameter $\Fg_\Pix^\T$. The proof of Lemma 
\ref{thm:convexity} is given in Appendix \ref{app:convexity}.
\end{lemma}

We also introduce a joint shape prototype vector $\bm{\Shape}:= [ 
\bm{\Shape}_1^\top \cdots  \bm{\Shape}_\NProt^\top ]^\top$ that is obtained by 
concatenating all individual shape prototype vectors $\bm{\Shape}_\Prot, \ \Prot 
\in \{1,\dots, \NProt\}$ (c.f. Fig. \ref{fig:model_shapes}). The component 
$\bm{\Shape}_{\Prot, \Pix^\prime}$ equals the negative logarithm of the 
background probability of pixel $\Pix^\prime$ in the normal shape prototype 
$\bm{\Shape}_\Prot$, 
\begin{equation} \label{eq:shape_vector}
  \bm{\Shape}_{\Prot, \Pix^\prime} = -\log \bigl( 1 - P_{\Prot} (\Fg 
_{\Pix^\prime} ^\T = 1) \bigr).
\end{equation}
The following lemma establishes a relationship between the foreground 
explanation term $J_\Pix ( \mathbf{\Lat} )$, the parsing indicator 
$\mathbf{\Lat}$ and the joint shape prototype vector $\bm{\Shape}$.
\pagebreak
\begin{lemma} \label{thm:pix_explanation}
The foreground explanation term $J_\Pix(\cdot)$ is the sum over all pixels 
$\Pix$ of convex functions $\Aux _{ \Fg_\Pix^\T} (\cdot) $ whose argument is 
a bilinear function of the parsing indicator $\mathbf{\Lat}$ and the joint 
shape prototype $\bm{\Shape}$,
\begin{alignat}{2} \label{eq:J_xi}
 J_\Pix ( \mathbf{\Lat} ) &=  \sum_\Pix \Aux_{\Fg_\Pix^\T} \bigl(
\bm{\Shape}^\top \mathbf{\MakeUppercase{\Xpar}}_\Pix \mathbf{\Lat} + \Xpar_0
\bigr).
\end{alignat}
The parameter matrices $\mathbf{ \MakeUppercase{\Xpar} }_\Pix$ and scalar 
$\Xpar_0$ do not depend on the parsing indicator $\mathbf{\Lat} $ or joint 
shape prototype $\bm{\Shape}$. The proof of Lemma \ref{thm:pix_explanation} is 
given in Appendix \ref{app:derivation_Jpix}.
\end{lemma}

In Lemmas \ref{thm:pix_explanation} and \ref{thm:hyp_explanation} we 
expressed the foreground and hypotheses explanation terms $J_\Pix(\cdot)$ and 
$J_\Hyp(\cdot)$ as convex functions of the parsing indicator $\mathbf{\Lat}$. 
Therefore, the video parsing objective function $J(\cdot) := J_\Pix(\cdot) + 
J_\Hyp(\cdot)$ (Eq. \ref{eq:J}) is a convex function of the parsing indicator 
$\mathbf{\Lat}$. To efficiently solve the optimization problem, we relax 
the parsing indicator $\mathbf{\Lat}$ to the positive simplex, 
$\mathbf{\Lat}_\Hyp \succeq 0$ and $ \mathbf{1}^\top \mathbf{\Lat}_\Hyp \leq 1, 
\ \forall \Hyp$. The last inequality follows from Eq. \ref{eq:subst} and the 
fact that $\Obj_\Hyp \leq 1$.

The MAP inference in our video parsing model is thus equivalent to 
the following constrained convex optimization problem,
\begin{alignat}{2} \nonumber 
  &\argmin_{\mathbf{\Lat}} J( \mathbf{\Lat} ) = \mathbf{\Hpar}^\top
\mathbf{\Lat} + \Hpar_0 + \sum_\Pix \Aux_{\Fg_\Pix^\T} \bigl( \bm{\Shape}^\top
\mathbf{ \MakeUppercase{\Xpar} }_\Pix \mathbf{\Lat} + \Xpar_0 \bigr), \\
\label{eq:objective}
& \text{s.t. }\mathbf{\Lat}_\Hyp \succeq 0 \text{ and } \mathbf{1}^\top
\mathbf{\Lat}_\Hyp \leq 1, \ \forall \Hyp.
\end{alignat}

After finding the optimal value of the parsing indicator $\mathbf{\Lat}$, we 
calculate the hypothesis indicator $\Obj_\Hyp$, and the matching normal object 
prototype $\Mod_\Hyp$ of hypothesis $\Hyp$, as 
\begin{alignat}{2} 
\label{eq:o_h}
 \Obj_\Hyp &= \sum_{\Prot=1}^\NProt \Lat_{\Hyp,\Prot}, \\ \label{eq:m_h}
 \Mod_\Hyp &= \argmax_{\Prot} \Lat_{\Hyp,\Prot}.
\end{alignat}

\subsection{Solving the Convex Optimization Problem} 

\begin{figure}[tp]
  \centering
  \includegraphics[width=0.4\textwidth,height=2in]{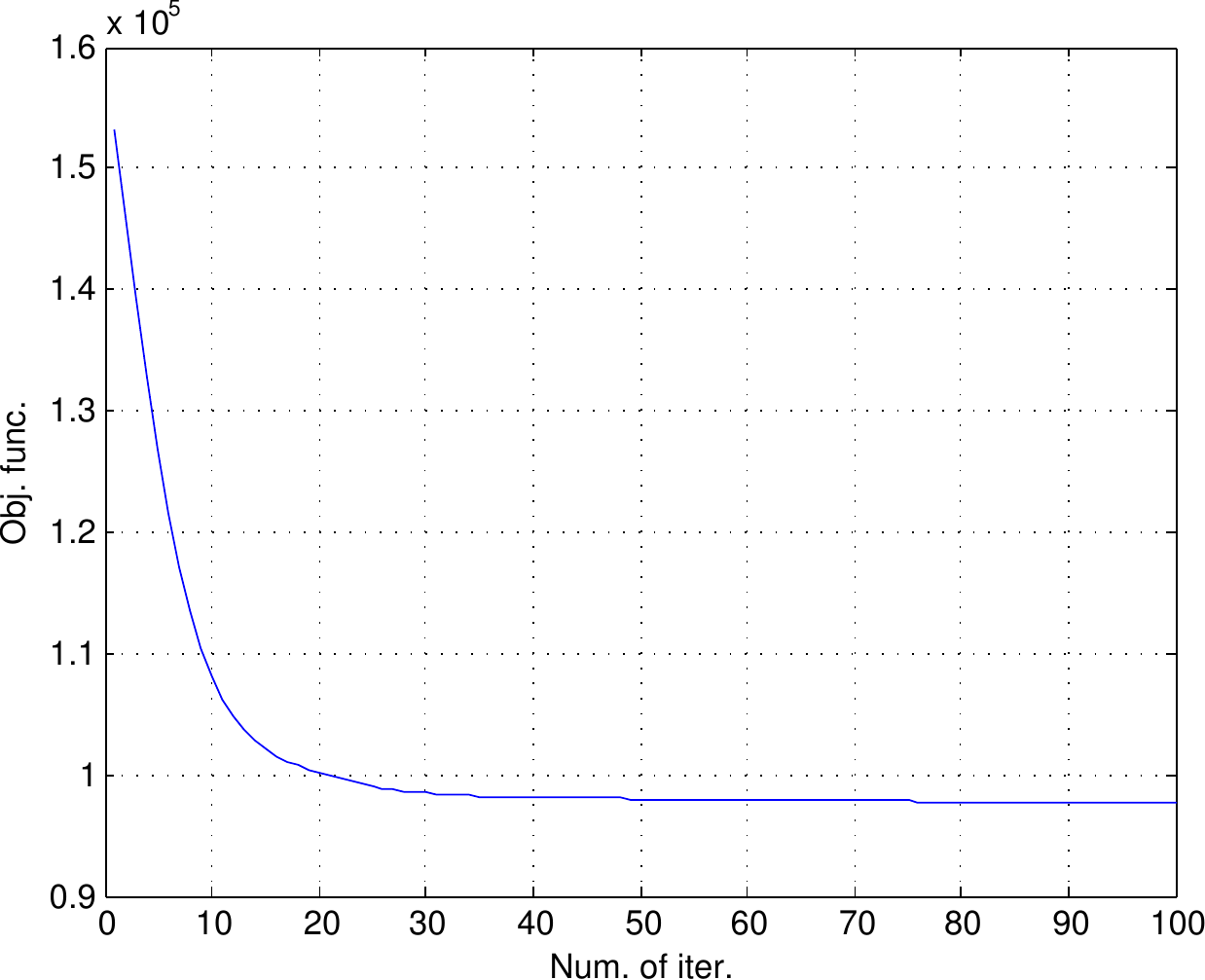}
\caption{Values of the objective function $J(\mathbf{\Lat})$ (Eq. 
\ref{eq:objective}) that are obtained as part of the convex optimization 
procedure that is used to solve the proposed video parsing problem. 
\label{fig:obj_func}}
\end{figure}

In the previous section, we showed that the joint inference of variables
$\{\Obj_\Hyp, \Mod_\Hyp\}_\Hyp$ can be achieved by minimizing the MAP objective
function $J(\mathbf{\Lat})$ to obtain the parsing indicator $\mathbf{\Lat}$ 
(Eq. \ref{eq:objective}), that belongs to the Cartesian product $ 
\mathbf{\MakeUppercase{\Lat}} = 
\mathbf{\MakeUppercase{\Lat}}_\Hyp \times \dots \times 
\mathbf{\MakeUppercase{\Lat}}_\Hyp $ of positive simplexes, 
\begin{equation} 
 \mathbf{\MakeUppercase{\Lat}}_\Hyp = \{\mathbf{\Lat}_\Hyp: \mathbf{\Lat}_\Hyp
\succeq 0 \text{ and } \mathbf{1}^\top \mathbf{\Lat}_\Hyp \leq 1 \}.
\end{equation}
The function $J(\mathbf{\Lat})$ is convex, smooth and bounded on the set 
$\mathbf{\MakeUppercase{\Lat}}$. 
The \textit{projected gradient} method \cite{Combettes11},
\begin{equation}  \label{eq:inf_gradient}
 \mathbf{\Lat}^{n+1} = \Proj_\mathbf{\MakeUppercase{\Lat}} (\mathbf{\Lat}^n -
\alpha_n \nabla_\mathbf{\Lat} J(\mathbf{\Lat}^n)), 
\end{equation}
converges to the global optimum of the convex optimization problem in Eq. 
\ref{eq:objective}, because of the Lipschitz-continuity of the first derivative 
of function $\Aux_{\Fg_\Pix^\T} (\cdot)$ as stated in the following lemma. 

\begin{lemma} \label{thm:lipschitz}
The first derivative of the function  $\Aux_{\Fg_\Pix^\T} (x) $ is  
$\rho$-Lipschitz continuous in argument $x \geq \Xpar_0$, i.e. 
there is a constant $\rho$ such that
\begin{equation}
 \bigl| \Aux_{\Fg_\Pix^\T}^\prime (x_1) - \Aux_{\Fg_\Pix^\T}^\prime (x_2) 
\bigr| \leq \rho | x_1 - x_2 |, \ \forall x_1, x_2 \geq \Xpar_0.
\end{equation}
The proof of Lemma \ref{thm:lipschitz} is given in Appendix 
\ref{app:lipschitz}.
\end{lemma}

The projection $\Proj_{\mathbf{\MakeUppercase{\Lat}}}(\cdot)$ requires each 
$\mathbf{\Lat}_\Hyp$ to be projected onto the positive simplex 
$\mathbf{\MakeUppercase{\Lat}}_\Hyp$. The projection onto the positive simplex 
is calculated by applying the method of Duchi et al. \cite{Duchi08}. The 
projected gradient method finds the solution of the video parsing after only 
few tens of iterations (Fig. \ref{fig:obj_func}). 

\subsection{From Inference to Abnormalities} \label{subsec:abnormality}

Video parsing analyses the foreground in a video and identifies objects that
have atypical appearance or behave suspiciously, to label these as abnormal.
Abnormalities can also be localized on the level of pixels, where it
leads to a segmentation of regions in the video that contain irregular 
spatio-temporal patterns. Subsequently, we see how  both the object-level and
pixel-level abnormalities can be detected in video, based on the inference
results of our video parsing approach.

\textbf{Object-level abnormalities}. A hypothesis $\Hyp$ is an abnormal object,
$\Abn_\Hyp = 1$, if it is indispensable for explaining the foreground, 
$\bar{\Obj}_\Hyp = 1$, but it does not have a matching normal object prototype, 
i.e., the best estimate $\bar{\Mod}_\Hyp$ of a matching prototype is unlikely to 
explain the hypothesis (cf. Eq. \ref{eq:matching}),
\begin{alignat}{2} \label{eq:objabn} \nonumber
&P(\Abn_\Hyp=1|\Obj_\Hyp = \bar{\Obj}_\Hyp, \Mod_\Hyp = \bar{\Mod}_\Hyp)  \\
&\propto \bar{o}_\Hyp P(\Obj_\Hyp=1 | \Dsc_\Hyp)  P(\Mod_\Hyp \neq 
\bar{m}_\Hyp | \Obj_\Hyp = \bar{o}_\Hyp, \Dsc_\Hyp, \Loc_\Hyp) \\
&\propto \bar{o}_\Hyp P(\Obj_\Hyp=1 | \Dsc_\Hyp) \Bigl( 1 - P(\Mod_\Hyp = 
\bar{m}_\Hyp | \Dsc_\Hyp) P(\Loc_\Hyp | \Mod_\Hyp = \bar{m}_\Hyp) \Bigr).
\end{alignat}

\textbf{Pixel-level abnormalities}. Similarly, a pixel $\Pix$ is part of an
abnormal object, $\Abn_\Pix^\T = 1$, if it is in the foreground, $\Fg _\Pix ^\T
= 1$, and at least one of the hypotheses that extend over this pixel, $\{h :
\Pix \in \mathcal{S}_\Hyp^\T\}$, is abnormal,
\begin{alignat}{2} \label{eq:pixabn} \nonumber
&P(\Abn_\Pix^\T=1 | \Fg_\Pix^\T, \{\Abn_\Hyp\}_{\Hyp : \Pix \in
\mathcal{S}_\Hyp^\T}) \\ 
&\propto \Fg_\Pix^\T \cdot P(\Fg_\Pix^\T = 1) \cdot \max_{\Hyp : \Pix \in
\mathcal{S}_\Hyp^\T} P(\Abn_\Hyp = 1 | \Obj_\Hyp, \Mod_\Hyp).   
\end{alignat}

\section{Learning an Object Model for Video Parsing} \label{sec:learning}

Parsing query videos for abnormality detection requires an object model.
We use training videos that contain a large number of normal object samples but 
no abnormalities to train the normal object model that consists of prototypes 
representing the normal object shape, appearance, and motion. As ground truth 
locations of objects in the training videos are not provided, we infer them by 
video parsing. However, for video parsing we need to know the normal object 
prototypes. A standard approach for solving such a problem of mutual 
dependencies is \textit{expectation-maximization} (EM) \cite{Dempster77}. Given 
an initial estimate of the normal object prototypes, we use them to parse the 
training videos, i.e. discover hypotheses that best explain the foreground and 
are matched to the object prototypes (E-step). Thereafter, we update the object 
prototypes using the matched hypotheses (M-step). We find the object model by 
iterating the EM steps until convergence.

The goal of learning is to estimate the normal object shape prototypes $ \{ 
\bm{\Shape} _\Prot \} _\Prot $ (Eq. \ref{eq:shape_vector}) and their 
corresponding spatio-temporal descriptors $\{ \Dsc_\Prot\}_\Prot$, $\Prot \in 
\{1, \dots, \NProt\}$.
The objective function for learning is the same as for the inference (Eq. 
\ref{eq:objective}), except that it is now minimized jointly in terms of 
shape prototypes $ \{ \bm{\Shape} _\Prot \} _\Prot$, their spatio-temporal 
descriptors $ \{ \Dsc_\Prot \}_\Prot $, as well as the parsing indicator 
$\mathbf{z}$ (Eq. \ref{eq:subst}),
\begin{alignat}{2} \nonumber
&\argmin_{ \{ \Dsc_\Prot, \bm{\Shape} _\Prot \} _\Prot, \mathbf{\Lat}}
J(\mathbf{\Lat}, \{ \Dsc_\Prot, \bm{\Shape} _\Prot \} _\Prot) =
J_\Hyp(\mathbf{\Lat}, \{ \Dsc_\Prot \}_\Prot) + J_\Pix(\mathbf{\Lat},
\{\bm{\Shape} _\Prot \} _\Prot), \\
\label{eq:learning}
&\text{ s.t. } \bm{\Shape}_\Prot \succeq 0, \ \forall \Prot, \ 
\mathbf{\Lat}_\Hyp \succeq 0 \text{ and } \mathbf{1}^\top \mathbf{\Lat}_\Hyp 
\leq 1, \ \forall \Hyp.
\end{alignat}

\begin{figure}[t] 
\centering
\includegraphics[width=3in]{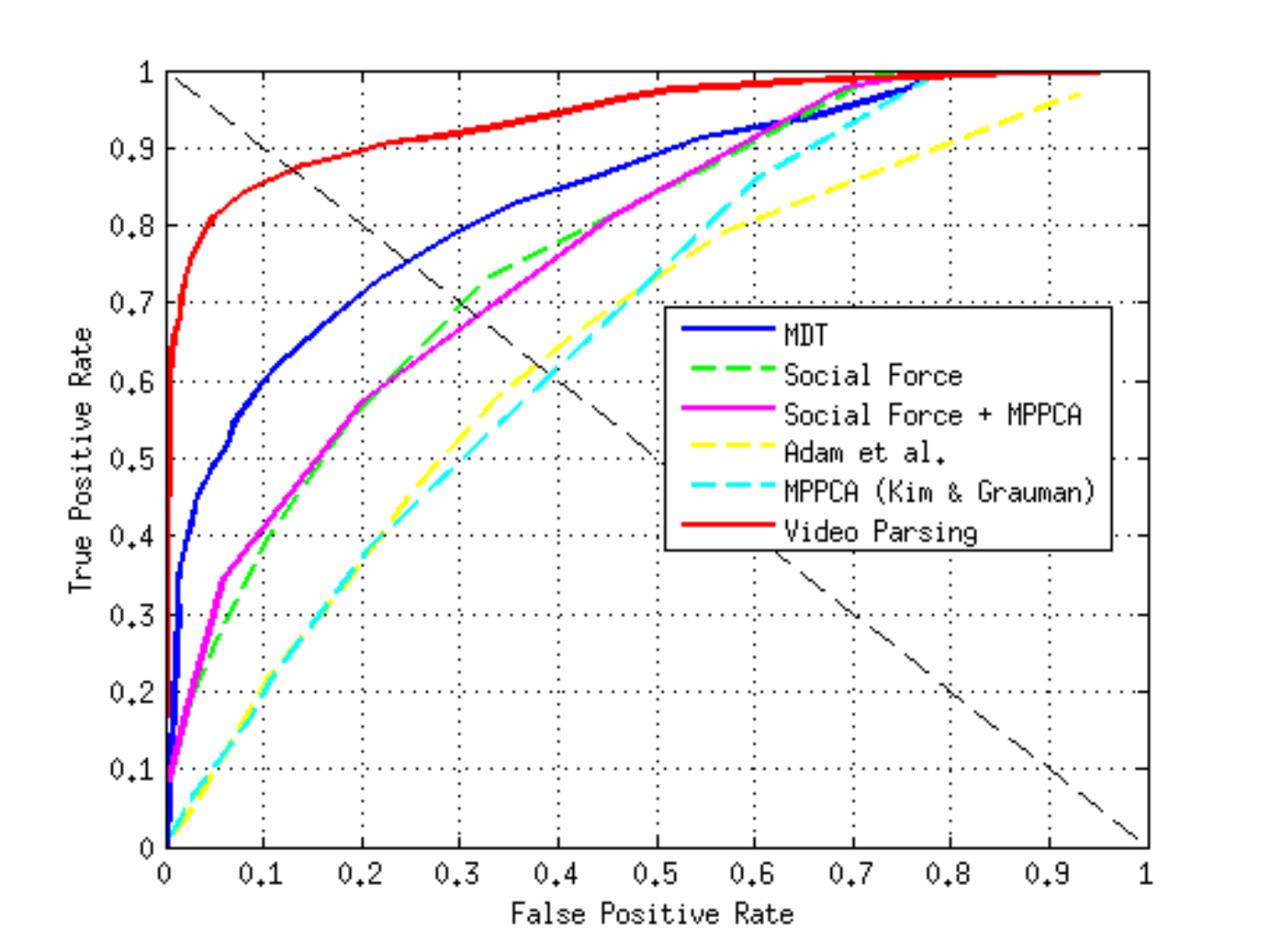}
\vspace{-0.2in}
\caption{Frame-wise abnormality labeling on the UCSD \textit{ped1} 
dataset. Performance measures AUC and EER given in Tab. \ref{tab:ped1tab} are 
calculated from the ROC curves. \label{fig:ped1roc}}
\end{figure}

The hypotheses explanation term $J_\Hyp(\cdot)$ is a function of the parsing 
indicator $\mathbf{\Lat}$ and the spatio-temporal descriptors $\{ \Dsc_\Prot 
\}_\Prot$,
\begin{equation} 
 J_\Hyp(\mathbf{\Lat}, \{ \Dsc_\Prot \}_\Prot) = \beta \sum_\Hyp \sum_\Prot
\Lat_{\Hyp, \Prot}  \Delta (\Dsc_\Hyp, \Dsc_\Prot) + \tilde{\mathbf{\Hpar}}
^\top \mathbf{\Lat} + \Hpar_0,
\end{equation}
where the parameters $\tilde{\mathbf{\Hpar}}$ and $\Hpar_0$ do not depend on 
the parsing indicator $\mathbf{\Lat}$ or the spatio-temporal descriptors $\{ 
\Dsc_\Prot \}_\Prot$ (see the proof of Lemma \ref{thm:hyp_explanation} in 
Appendix \ref{app:derivation_Jh}). 

From Eq. \ref{eq:J_xi} we see that the foreground explanation term 
$J_\Pix(\cdot)$ depends in a convex way on both the parsing indicator 
$\mathbf{\Lat}$ and the joint shape prototype vector $\bm{\Shape}$.

\textbf{Procedure for the object prototype learning}. We now explain the EM 
algorithm used for solving the optimization problem of Eq. \ref{eq:learning}: 

\textbf{E-step}. Given the object prototypes, we parse the training videos to 
infer the parsing indicator $\mathbf{\Lat}$ (Eq. \ref{eq:objective}) that 
yields the hypothesis indicator $\Obj_\Hyp$ for each hypothesis $\Hyp$, and its 
corresponding normal object prototype $\Mod_\Hyp$ (Eq. \ref{eq:o_h} and 
\ref{eq:m_h}).

\textbf{M-step}. We estimate the shape prototypes $\{ \bm{\Shape} _\Prot \} 
_\Prot$ and their spatio-temporal descriptors $\{ \Dsc_\Prot \}_\Prot$ from 
the results of video parsing. As hypotheses overlap in training videos, the 
corresponding shape prototypes become mutually dependent and thus need to 
be learned jointly. We estimate the joint shape prototype vector $\bm{\Shape}$ 
by the following convex optimization,
\begin{equation} \label{eq:shape_opt}
 \bm{\Shape} = \argmin_{\tilde{\bm{\Shape}} \succeq 0}
J_\Pix(\mathbf{\Lat},\tilde{\bm{\Shape}}) = \sum_{\Pix} \Aux_{\Fg_\Pix^\T}
( \tilde{\bm{\Shape}}^\top \mathbf{ \MakeUppercase{\Xpar} }_\Pix \mathbf{\Lat} +
\Xpar_0). 
\end{equation}
 The convex optimization problem of Eq. \ref{eq:shape_opt} can be solved 
efficiently by the projected gradient method that we used for solving the MAP
inference problem (Eq. \ref{eq:inf_gradient}), 
\begin{equation} 
 \bm{\Shape}^{n+1} = \Proj_{\mathbb{R}_+^{|\Shape|}} (\bm{\Shape}^n - \alpha_n
\nabla_{\bm{\Shape}} J_\Pix(\mathbf{\Lat}, \bm{\Shape}^n)). 
\end{equation}
The spatio-temporal descriptors $\{\Dsc_\Prot\}_\Prot, \ \Prot 
\in \{1, \dots, \NProt \}$ are estimated separately for each normal object 
prototype, 
\begin{equation}
 \Dsc_\Prot = \argmin_{\tilde{\Dsc}_\Prot} \sum_\Hyp \Lat_{\Hyp,\Prot} 
\Delta(\Dsc_\Hyp, \tilde{\Dsc}_\Prot).  
\end{equation}
In case of a squared Euclidean distance function, $\Delta ( \Dsc_\Hyp, 
\Dsc_\Prot ) = \| \Dsc_\Hyp - \Dsc_\Prot \|^2$, there is a closed-form solution 
for $\Dsc_\Prot$, given as an average of spatio-temporal descriptors 
$\Dsc_\Hyp$ of those hypotheses that are matched to prototype $\Prot$ by 
video parsing,
\begin{equation}
 \Dsc_\Prot = \frac{\sum_\Hyp  \Lat_{\Hyp,\Prot} \Dsc_\Hyp}{ \sum_{\Hyp}
\Lat_{\Hyp,\Prot} }.
\end{equation}
The EM algorithm assumes uniform location and velocity distributions (Eq. 
\ref{eq:parzen}) for normal object prototypes. However, after the EM algorithm 
is converged, we estimate the prototype's location and velocity distributions 
from matched object hypotheses by the non-parametric \textit{Parzen} windows.

\begin{table} 
  \caption{Performance measures on the UCSD \textit{ped1} dataset 
\label{tab:ped1tab}}
    \begin{tabular}{|l||c|c||c|c||c|c|}
    \hline
    ~ & \multicolumn{2}{c||}{frame-wise} & 
\multicolumn{2}{c||}{\parbox{1.4cm}{\centering pixel-wise \\ partial }} & 
\multicolumn{2}{c|}{\parbox{1.4cm}{\centering pixel-wise \\ full }} \\ 
\hline   
    ~  & \parbox{0.6cm}{\centering AUC \\ ($\%$)} & 
\parbox{0.5cm}{\centering EER \\ ($\%$)} & \parbox{0.5cm}{\centering AUC \\ 
($\%$)} & \parbox{0.5cm}{\centering RD \\ ($\%$)} & \parbox{0.5cm}{\centering 
AUC \\ ($\%$)} & \parbox{0.5cm}{\centering RD \\ ($\%$)} \\ \hline \hline
    Social force \cite{Mehran09} & 67.5 & 31 & 19.7 & 21 & -  & -  \\ \hline
    MPPCA \cite{Kim09} & 59 & 40  & 20.5  & 18 & - & -  \\ \hline
    \parbox{1.5cm}{Social force \\ + MPPCA} & 67 & 32  & 21.3 & 28 & - & -  \\ 
\hline
    Adam \cite{Adam08} & 65 & 38  & 13.3  & 24 & - & -  \\ \hline
    Sparse \cite{Cong11} & 86 & 19  & 46.1  & 46 & - & -  \\ \hline
    LSA \cite{Saligrama12} & 92.7 & 16  & -  & - & - & -  \\ \hline
    SCL \cite{Lu13} & 91.8 & 15  & 63.8  & 59.1 & - & -  \\ \hline
    MDT \cite{Mahadevan10} & 81.8 & 25  & 44.1  & 45 & - & -  \\ \hline
    HMDT CRF \cite{Li13} & - & 17.8  & 66.2  & 64.8 & 82.7 & 74.5  \\ \hline
    \textit{SVP} \cite{AnticO11} & 91 & 18  & 75.6  & 68 & 83.6 & 77  
\\ \hline
    \textit{STVP}   & \textbf{93.9} & \textbf{12.9}  & \textbf{80.3}  & 
\textbf{75.2} & \textbf{84.2} & \textbf{79.5}  \\ \hline
    \end{tabular}
\end{table}

\begin{figure*}[t] 
\begin{center}
\subfloat[]{ \label{fig:frameLabeling}
\includegraphics[width=.31\textwidth]{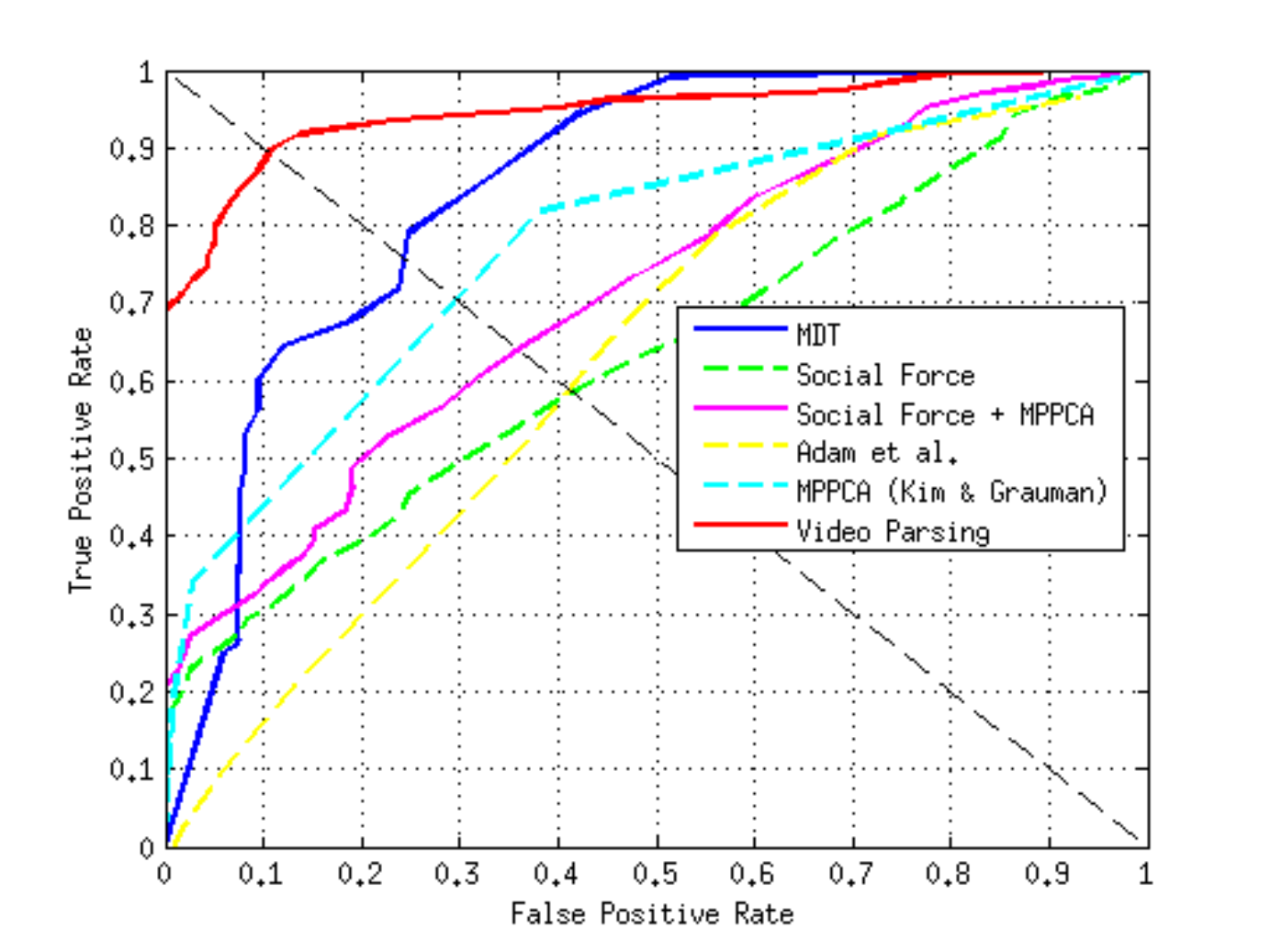}
}
\hfill
\subfloat[]{ \label{fig:pixelLabeling}
\includegraphics[width=.31\textwidth]{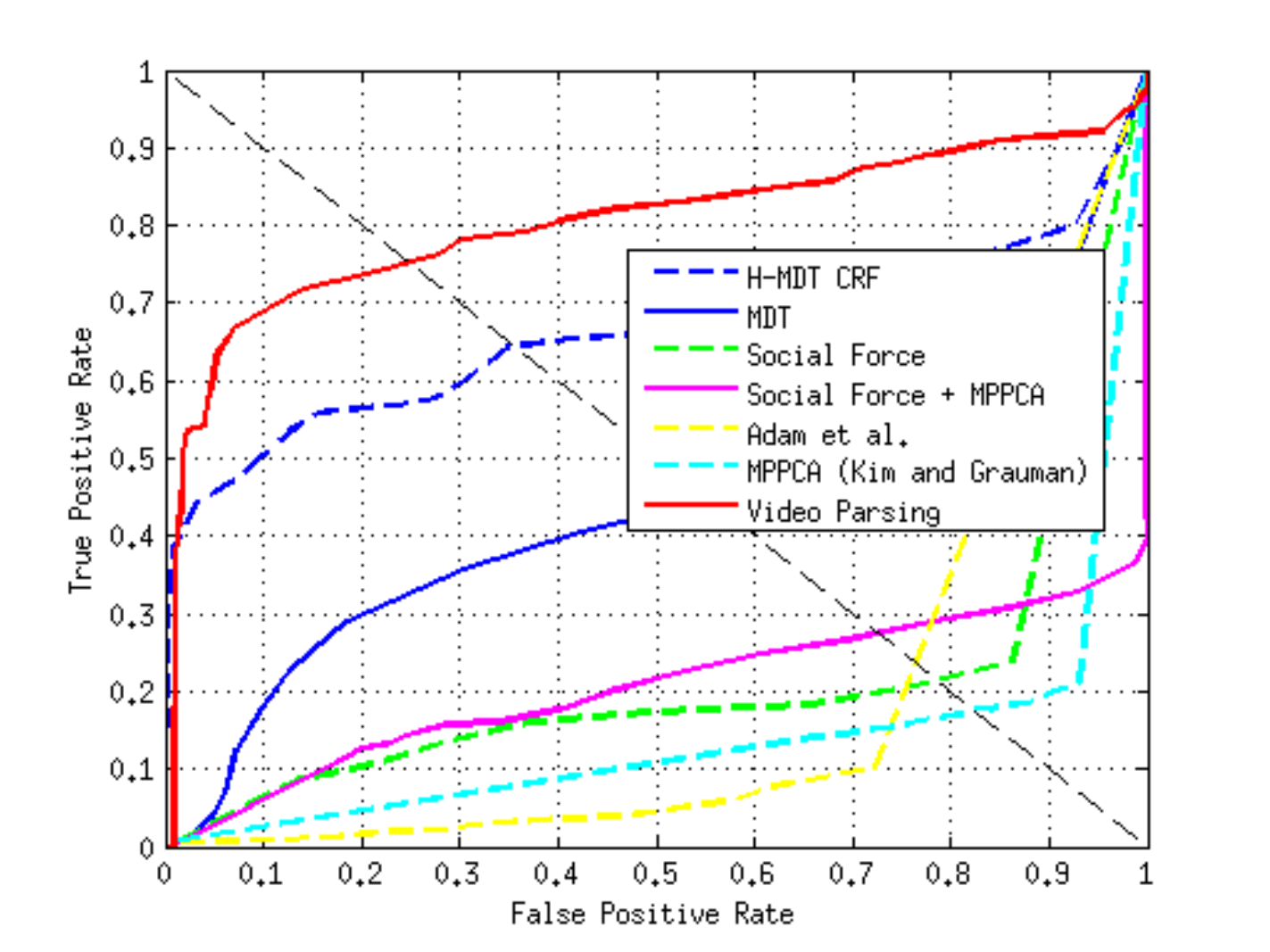}
}
\hfill
\subfloat[]{ \label{fig:pixelLabeling}
\includegraphics[width=.31\textwidth]{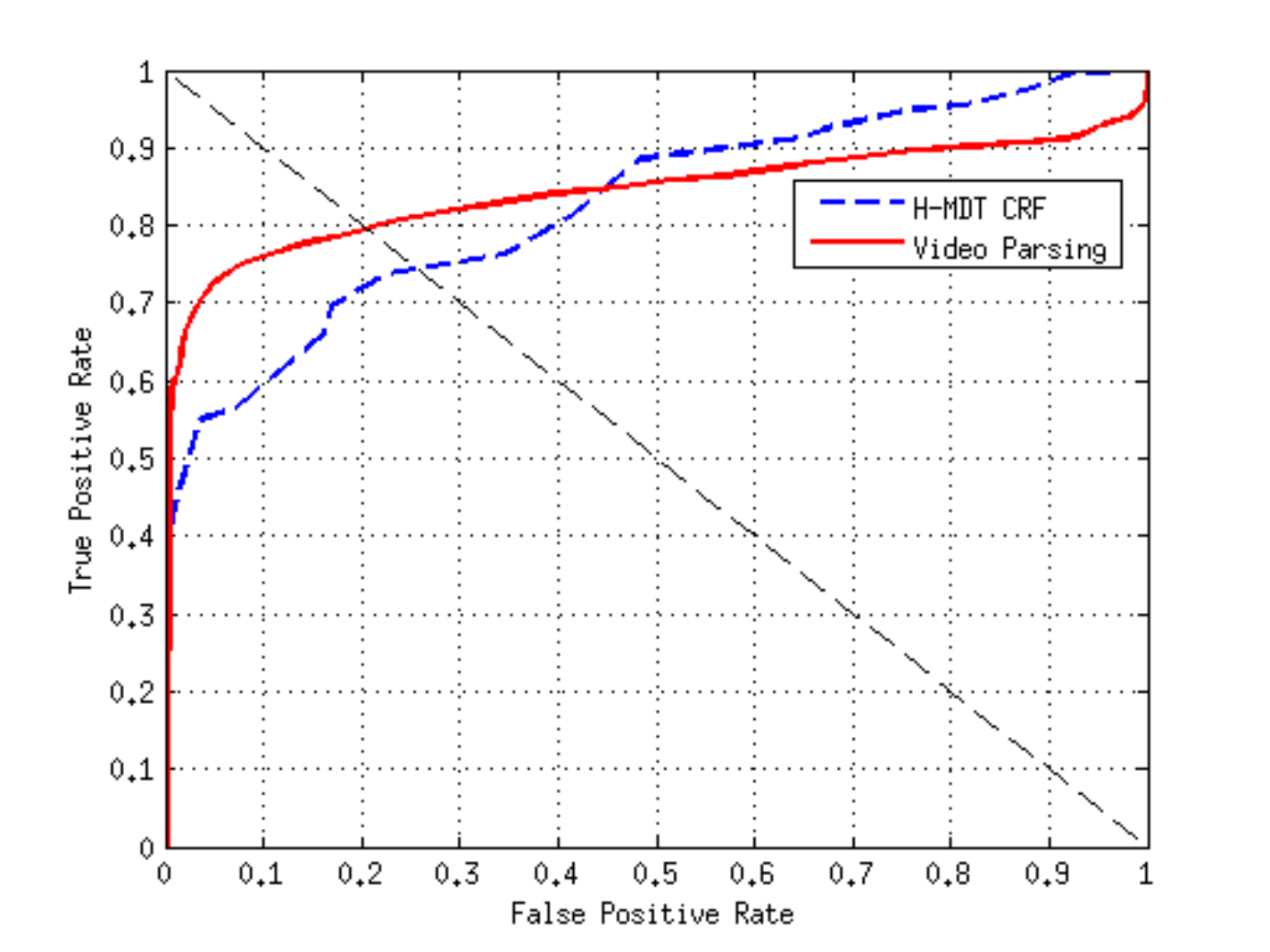}
}
\end{center}
\caption{(a) Frame-wise abnormality labeling for the UCSD \textit{ped2} 
dataset. (b) Pixel-wise abnormality prediction evaluated by the partially 
annotated UCSD \textit{ped1} dataset. (c) Pixel-wise abnormality prediction 
that is evaluated using the full annotation of the complete  UCSD \textit{ped1} 
dataset that we have assembled. In all of these cases our approach 
significantly 
improves upon the state-of-the-art, which can also be seen from the 
corresponding AUC and RD values provided in Tab. \ref{tab:ped1tab} and 
\ref{tab:ped2tab}. \label{fig:ped2roc}}
\end{figure*}

\begin{figure*}[t] 
\begin{center}$
\begin{array}{cccc}
\includegraphics[width=0.22\textwidth]
{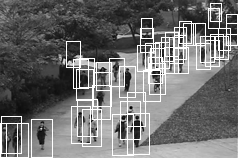} &
\includegraphics[width=0.22\textwidth]
{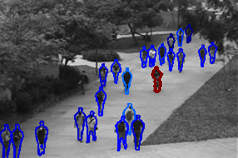} &
\setlength\fboxsep{0pt}
\fbox{\includegraphics[width=0.22\textwidth]
{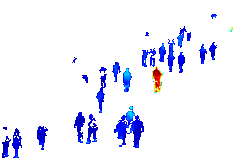}} &
\includegraphics[width=0.22\textwidth]
{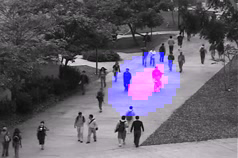} \\
\includegraphics[width=0.22\textwidth]
{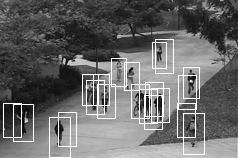} &
\includegraphics[width=0.22\textwidth]
{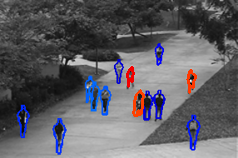} &
\setlength\fboxsep{0pt}
\fbox{\includegraphics[width=0.22\textwidth]
{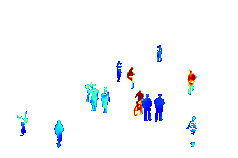}} &
\includegraphics[width=0.22\textwidth]
{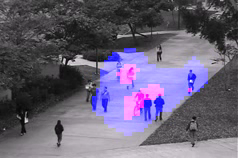} \\
\includegraphics[width=0.22\textwidth]
{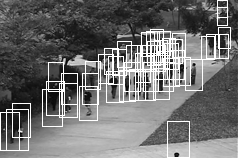} &
\includegraphics[width=0.22\textwidth]
{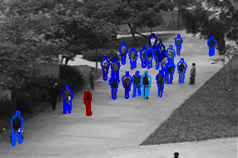} &
\setlength\fboxsep{0pt}
\fbox{\includegraphics[width=0.22\textwidth]
{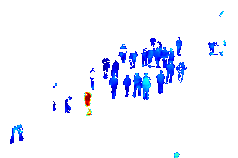}} &
\includegraphics[width=0.22\textwidth]
{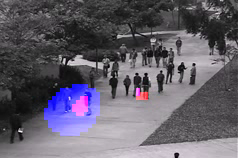} \\
\includegraphics[width=0.22\textwidth]
{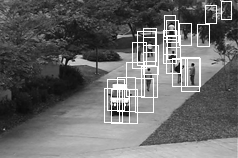} &
\includegraphics[width=0.22\textwidth]
{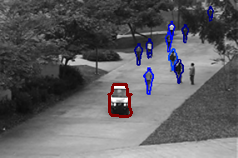} &
\setlength\fboxsep{0pt}
\fbox{\includegraphics[width=0.22\textwidth]
{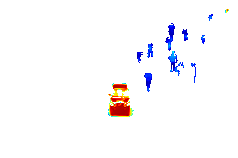}} &
\includegraphics[width=0.22\textwidth]
{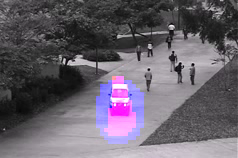}
\end{array}$
\vfill
\includegraphics[width=0.4\textwidth,height=0.3in]{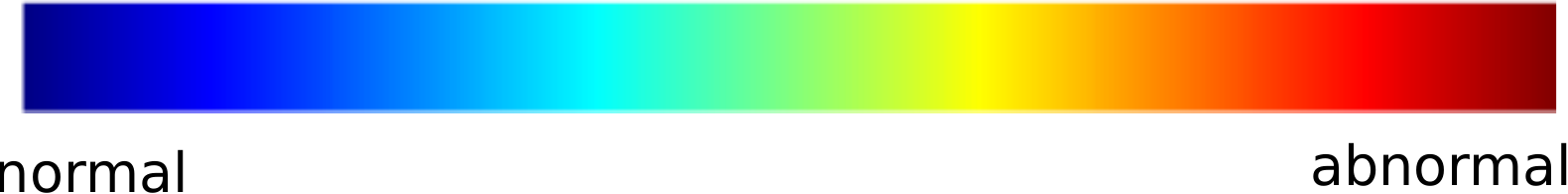}
\end{center}
\caption{Rows show results on different frames of the UCSD \textit{ped1} 
benchmark. Column i) the initialization of the video parsing by a shortlist 
of object hypotheses, column ii) hypotheses selected by video parsing with the 
best matching shape prototype colored according to abnormality probability 
$P(\Abn_\Hyp^\T=1)$, column iii) foreground pixel abnormality probabilities 
$P(\Abn_\Pix^\T=1)$, column iv) results by the H-MDT CRF approach \cite{Li13}. 
Best viewed in color. \label{fig:examplesNuno}} 
\end{figure*}

\textbf{Initialization}. To start the EM algorithm, we need an initial estimate 
of the normal object model. After background subtraction, some foreground 
segments correspond to isolated normal objects that can be used to initialize 
our object prototypes. However, foreground/background segmentation produces also 
many foreground segments which correspond to interacting objects (doublets, 
triplets etc.). These segments are more complex and can be analyzed only by 
video parsing. Consequently, we need to infer which of the training foreground 
segments correspond to isolated normal objects and estimate object prototypes 
based upon them. We observe that isolated normal objects create compact clusters 
in the feature space. On the other hand, segments that are mixtures of two or 
more objects are diverse and spread out in the feature space. To detect isolated 
normal objects, we cluster all the foreground segments and then select compact 
clusters in the feature space that correspond to isolated objects. We use 
\textit{Ward's} method for agglomerative clustering to minimize the variance 
of clusters. Normal object prototypes are then computed as the centers of 
compact clusters.

\section{Creating Initial Object Hypotheses} \label{sec:shortlist}

To initialize video parsing, we need a shortlist of spatio-temporal object
hypotheses $\Hyp$ (Sect. \ref{sec:parsing_model}). A spatio-temporal hypothesis
$\Hyp$ consists of a sequence of object candidates in individual frames that are
linked temporally. In this section we explain a method for producing per-frame 
object candidates and group them temporally based on their motion to 
obtain the shortlist of Sect. \ref{sec:parsing_model}. Thereafter, we explain 
how to fill-in per-frame candidates that were missed during temporal grouping.

\textbf{Temporal grouping of per-frame object candidates}. To detect per-frame 
object candidates, we apply an inverted background detector that is trained to 
distinguish background patterns from everything else. The inverted background 
detector is trained on background and normal foreground segments obtained from 
training videos by background subtraction. The discriminative appearance-based 
classifier retains in each frame the object candidates that are least likely to 
be background. The standard non-maximum suppression (NMS) then removes some of 
the candidates based on the overlap criteria. The discriminative classifier is 
trained using a linear SVM \cite{Chang11} with frame-wise descriptor of 
Eq. \ref{eq:descriptor} extracted from background/foreground segments of 
training videos.

We then employ agglomerative clustering to perform a temporal grouping of
candidates. This yields spatio-temporal hypotheses $\Hyp$, which are
sequences of per-frame candidates. As usual, the clustering starts with
singleton clusters (each candidate being a cluster). Then, in each round of the
recursive clustering, those groups of per-frame object candidates which are 
most similar based on their motion and which do not share the same frames are 
grouped. The motion of a candidate is represented by the set of trajectories 
obtained by tracking the edge points inside the support region of a candidate. 
For tracking the feature points we use optical flow vectors that are previously 
computed by the method of \cite{Liu08}. We now define similarity of two object 
candidates as the ratio of the number of feature point trajectories that are 
shared by two candidates over the total number of trajectories in two 
candidates.  As the result of temporal grouping, we obtain a shortlist of 
spatio-temporal hypotheses $\Hyp$.

\textbf{Filling-in missing candidates by Kalman filter}. The inverted 
background detector used for producing object candidates in each frame 
typically has a number of missed detections. These are the frames in which 
none of the object candidates is associated with a hypothesis $\Hyp$. We 
fill-in the missed object detections with the contextual help of other 
per-frame candidates that belong to the same hypothesis $\Hyp$. Therefore, the 
location of a missed object candidate $\Loc_\Hyp^\T$ at time $\T$ is estimated 
from the available object candidate locations at times $ \{ \T_1, \T_2, \dots \} 
$ by a non-causal Kalman filter.

The shortlist of object hypotheses established by temporal grouping has a high
recall at the cost of low precision. By maximizing the recall, the  shortlist 
includes all relevant hypotheses, while still maintaining a reasonable 
total number thereof (about one hundred). Since hypotheses are created by 
bottom-up grouping, there will, however, be many spurious hypotheses that can 
only be eliminated by video parsing.

\section{Experimental Evaluation} \label{sec:experiments}

We use three standard state-of-the-art benchmark sets for evaluating our video 
parsing approach and comparing its performance to the other state-of-the-art 
methods. We first analyze the detection results of our approach on the UCSD 
benchmark sets \textit{ped1} and \textit{ped2}, then we present additional 
results on the UMN benchmark set. We apply the standard evaluation protocol of 
the datasets.

\subsection{Evaluation on the UCSD Anomaly Datasets}

\subsubsection{Datasets Description}

We use the challenging UCSD anomaly datasets \textit{ped1} and \textit{ped2}, 
that were recently proposed by Mahadevan et al. \cite{Mahadevan10} for measuring 
the performance of abnormality detection algorithms. Both datasets consist 
of videos recorded in crowded walkway scenes that also feature lots 
of challenging abnormal instances which are objects with unusual appearance 
or behavior. The UCSD \textit{ped1} set contains $34$ training and $36$ test 
videos that are all $200$ frames long. Due to the low resolution of 
\textit{ped1} videos, the pedestrians who walk towards and away from the camera 
are only $10-25$ pixels high. In the UCSD \textit{ped2} dataset there are $16$ 
training and $12$ test videos that have a variable length (at most $180$ 
frames). Pedestrians in these videos are about $30$ pixels  high. Videos from 
both benchmark sets are very crowded, so that object heavily occlude one 
another.

Abnormalities in the UCSD datasets are not staged but occur naturally in the 
scene and can be grouped into: i) objects that do not fit to the context of 
the scene, such as a car on a crowded walkway, or ii) objects 
that look normal but behave in unusual way, such as people that cycle or 
skateboard across the walkway or walk in the lawn. Abnormalities from the 
UCSD benchmark sets include also carts and wheelchairs. We emphasize that the 
training videos consist only of normal objects and actions, so that a model for 
abnormalities cannot be learned from it.

\subsubsection{Evaluation Protocol}

We use the standard protocol for evaluating abnormality detection results that 
was proposed by Mahadevan et al. \cite{Mahadevan10}. The protocol consists of 
frame-wise and pixel-wise criteria. The frame-wise criterion labels a frame as 
abnormal if it contains at least one abnormal object detection. The localization 
accuracy of detected abnormalities is verified by the pixel-wise criterion that 
is more rigorous than the frame-wise criterion, since the detected abnormalities 
are compared to a pixel-level ground-truth mask. The pixel-wise criterion 
requires that at least $40\%$ of all ground-truth abnormal pixels to be marked 
as abnormal in order to count a frame as true positive. By calculating the true 
positive rate (TPR) and false positive rate (FPR) at different detection 
thresholds we obtain the receiver operating characteristic (ROC). 

Frame-wise and pixel-wise criteria use the area under the curve (AUC) as a 
performance measure calculated directly from the corresponding ROC curve. For 
the frame-wise criterion we calculate also the equal error rate (EER) as a value 
obtained when the false positive and false negative rates are equal. For 
pixel-wise criterion we compute the rate of detection (RD), that is equal to 
$1-$EER. The pixel-wise criterion is applied on the partially labeled UCSD 
\textit{ped1} dataset originally provided with the pixel-wise ground-truth 
annotation. Moreover, we also provide complete pixel-wise ground-truth 
annotations for the full datasets and evaluate thereon.

\begin{table}
  \renewcommand{\arraystretch}{1.3}
  \caption{Performance measures on the UCSD ped2 dataset 
\label{tab:ped2tab}}
  \centering
  \begin{tabular}{|l||c|c||c|c|}
    \hline
    ~ & \multicolumn{2}{c||}{frame-wise} &
\multicolumn{2}{c|}{\parbox{1.8cm}{\centering pixel-wise}} \\ \hline
    ~  & \parbox{0.6cm}{\centering AUC \\ ($\%$)} & \parbox{0.6cm}{\centering 
EER \\ ($\%$)} & \parbox{0.6cm}{\centering AUC \\ ($\%$)} & 
\parbox{0.6cm}{\centering RD \\ ($\%$)} \\ \hline \hline
    Social force \cite{Mehran09} & 63 & 42 & -  & -  \\ \hline
    MPPCA \cite{Kim09} & 77 & 30  & - & -  \\ \hline
    \parbox{2.8cm}{Social force + MPPCA}  & 71 & 36 & - & -  \\ \hline
    Adam \cite{Adam08} & 63 & 42  & - & -  \\ \hline
    MDT \cite{Mahadevan10} & 85 & 25  & - & -  \\ \hline
    H-MDT CRF \cite{Li13} & - & 18.5  & -  & 70.1 \\ \hline
    \textit{SVP} \cite{AnticO11} & 92 & 14  & -  & - \\ \hline
    \textit{STVP} & \textbf{94.6} & \textbf{10.6}  & \textbf{81.1}  & 
\textbf{78.8} \\ \hline
    \end{tabular}
\end{table}

\subsubsection{The Results of Evaluation}

Fig.\ \ref{fig:examplesNuno} compares the abnormality localization of our 
video parsing to the H-MDT CRF method \cite{Li13} on UCSD \textit{ped1} test 
videos. The first row shows a person riding a bike in a 
group of walking persons. In the second row there are three abnormalities in 
the scene: a person riding a bike, and two persons running along the walkway. 
The third row shows a person skateboarding along the walkway, and the fourth 
row shows an unusual object (car) in the scene. The columns show: (i) initial 
hypotheses of video parsing, (ii) hypotheses selected by video parsing, (iii) 
abnormality localization results of video parsing, (iv) abnormality 
localization results of H-MDT CRF method \cite{Li13}. Due to our learned 
normal shape model used for explaining the foreground, we achieve better 
localization of the abnormalities in videos.  

\begin{figure*}
\centering
\begin{minipage}[t]{.7\textwidth}
\begin{center}
\subfloat[]{ \label{fig:firstfig} 
\includegraphics[width=0.5\textwidth]{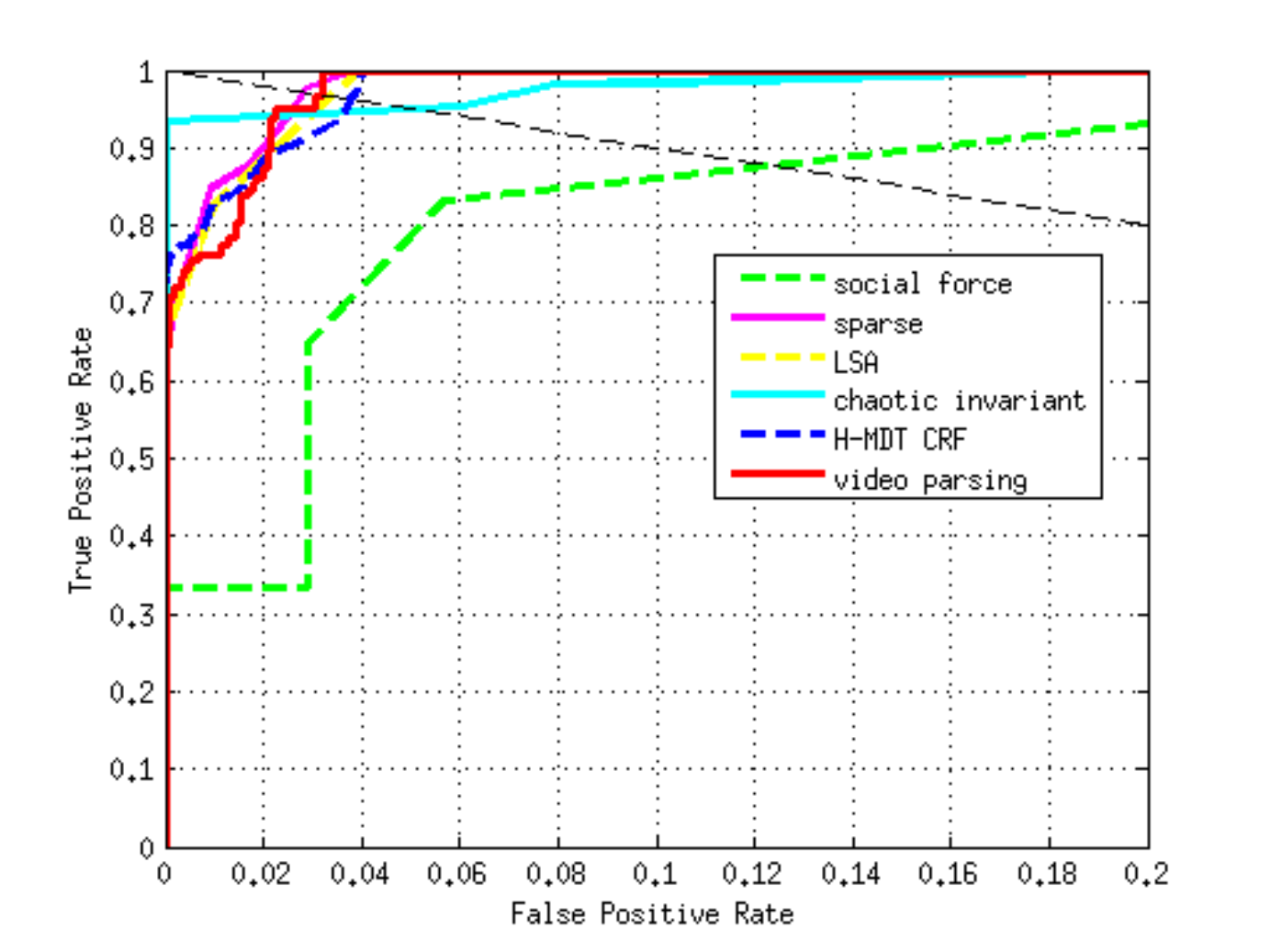}
}
\hfill
\subfloat[]{ \label{fig:secondfig}
\includegraphics[width=0.45\textwidth]{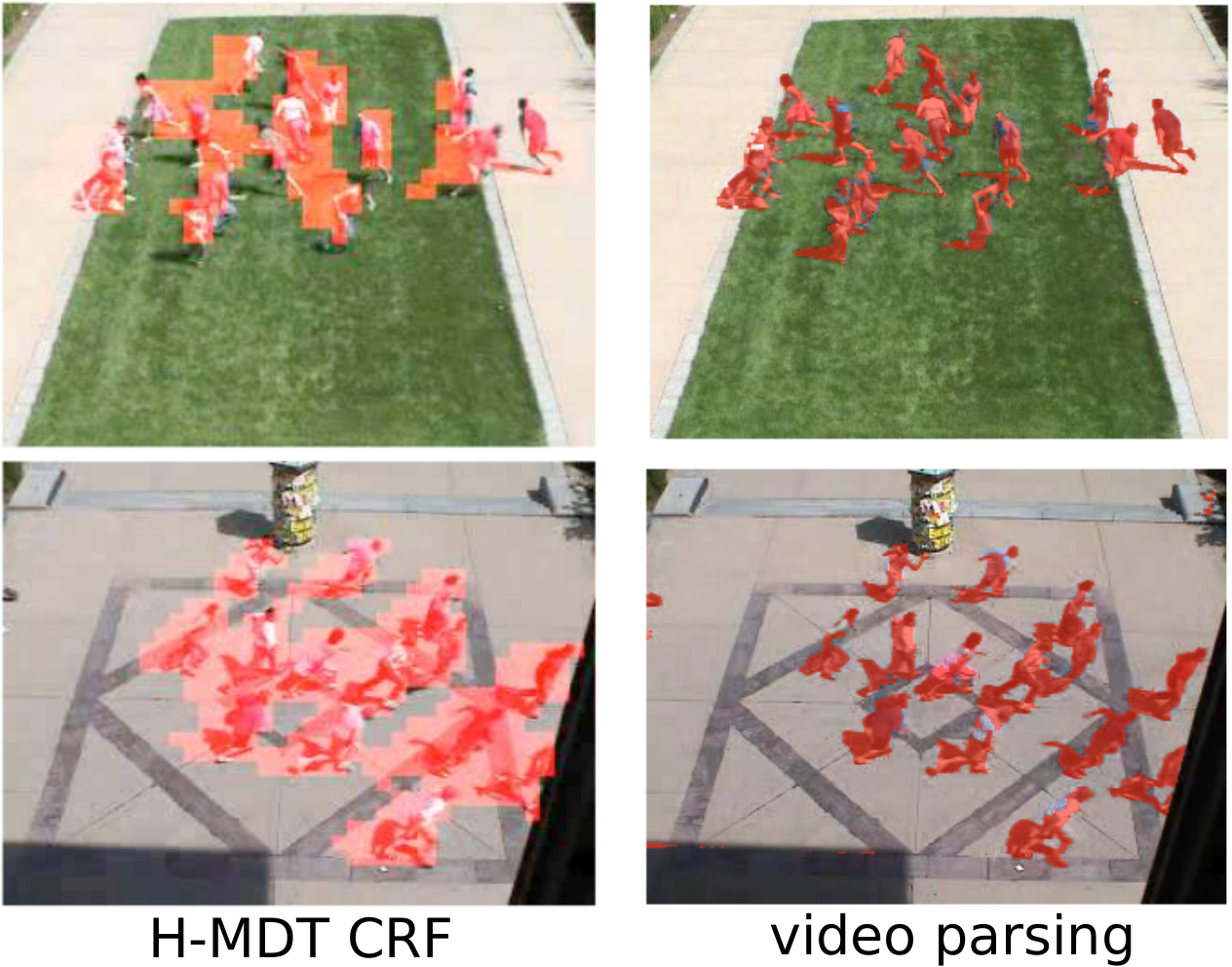}
}
\end{center}
\vspace{-0.15in}
\caption{Abnormality detection on the UMN dataset. (a) ROC curves for 
frame-wise labeling. (b) Detection results of the H-MDT CRF \cite{Li13} (left 
column) and video parsing (right column). Our approach exhibits competitive 
performance as can also be seen from the corresponding AUC and EER statistics 
in 
Tab. \ref{tab:umn_quant}. \label{fig:umnframe}}
\end{minipage}\hfill
\begin{minipage}[t]{.3\textwidth}
\centering
\vspace{-0.0in}
\captionof{table}{Performance measures on \newline the UMN dataset 
\label{tab:umn_quant}}
\small
\centering
\begin{tabular}{|p{0.55\linewidth}|p{0.55cm}|p{0.55cm}|}
\hline
method  & \parbox{0.55cm}{\centering AUC \\($\%$)} & \parbox{0.55cm}{\centering 
EER \\($\%$)} \\ \hline \hline
chaotic invariants \cite{Wu10}  & 99.4 &  5.3 \\ \hline
social force \cite{Mehran09} & 94.9 &  12.6 \\ \hline
LSA \cite{Saligrama12} & \textbf{99.5} &  3.4 \\ \hline
H-MDT CRF \cite{Li13} & \textbf{99.5} & 3.7 \\ \hline
Sparse \cite{Cong11} (scene1)  & \textbf{99.5} & - \\
Sparse \cite{Cong11} (scene2) & \textbf{97.5} & - \\
Sparse \cite{Cong11} (scene3) & 96.4 & - \\ \hline
\textit{STVP} (scene1) & \textbf{99.5} & 3.2 \\
\textit{STVP} (scene2) & \textbf{97.5} & 6.2 \\
\textit{STVP} (scene3) & \textbf{99.9} & \textbf{1.5} \\
\hline
\end{tabular}
\end{minipage}
\end{figure*}

\begin{figure*}[t]
\begin{center}$
\begin{array}{cccc}
\includegraphics[width=0.21\textwidth]
{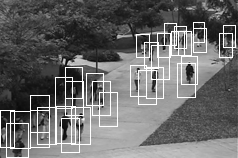} &
\includegraphics[width=0.21\textwidth]
{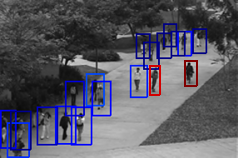} &
\includegraphics[width=0.21\textwidth]
{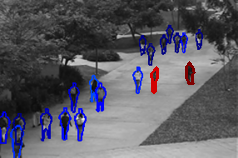} &
\setlength\fboxsep{0pt}
\fbox{\includegraphics[width=0.21\textwidth]
{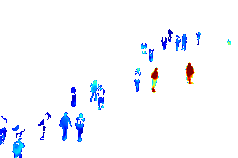}} \\
\includegraphics[width=0.21\textwidth]
{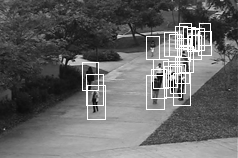} &
\includegraphics[width=0.21\textwidth]
{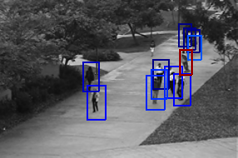} &
\includegraphics[width=0.21\textwidth]
{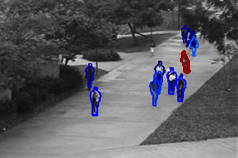} &
\setlength\fboxsep{0pt}
\fbox{\includegraphics[width=0.21\textwidth]
{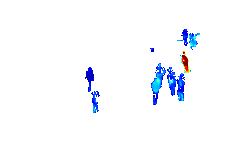}} \\
\includegraphics[width=0.21\textwidth]
{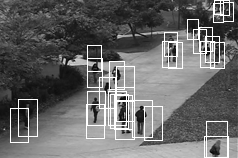} &
\includegraphics[width=0.21\textwidth]
{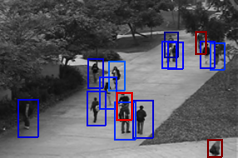} &
\includegraphics[width=0.21\textwidth]
{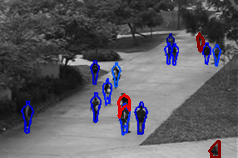} &
\setlength\fboxsep{0pt}
\fbox{\includegraphics[width=0.21\textwidth]
{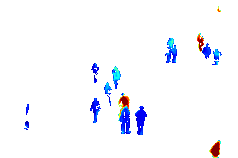}}
\end{array}$
\includegraphics[width=0.4\textwidth,height=0.3in]{./figures/colorbar}
\end{center}
\caption{Additional results of video parsing on the UCSD \textit{ped1} dataset. 
Rows correspond to different examples. The first, third and fourth column 
correspond to the first three columns of Fig. \ref{fig:examplesNuno}. The 
second column shows hypotheses that are selected from the shortlist by video 
parsing. Other hypotheses are discarded by explaining away using the selected 
hypotheses. Best viewed in color. \label{fig:additinal_examples}}
\end{figure*}

In Fig. \ref{fig:additinal_examples} we show more examples of the video 
parsing on UCSD \textit{ped1} test videos. Row $1$ shows two persons 
skateboarding and cycling on a very crowded walkway, row $2$ a skateboarder 
in a group of pedestrians, and row $3$ two cyclists and a person walking across 
the walkway. By comparing the first two columns one can see that most 
hypotheses from the shortlist are discarded by video parsing because they get 
statistically explained away. 

We also compare quantitatively our video parsing approach to the  
state-of-the-art methods on the challenging UCSD \textit{ped1} and \textit{ped2} 
benchmarks \cite{Mahadevan10}. The methods used in our comparison are the 
mixture of dynamic textures (MDT) \cite{Mahadevan10}, H-MDT CRF \cite{Li13}, 
social force model (SF) \cite{Mehran09}, mixture of optical flow (MPPCA) 
\cite{Kim09}, optical flow method (Adam et al.) \cite{Adam08}, SF+MPPCA 
\cite{Mahadevan10}, sparse reconstruction (Sparse), local statistical 
aggregates (LSA) \cite{Saligrama12}, and sparse combination learning 
(SCL) \cite{Lu13}. Our previous approach \cite{AnticO11} which parses video 
frames individually, one after another, is denoted as sequential video parsing 
(SVP). We denote by STVP the full spatio-temporal video parsing proposed in this 
paper.    

\begin{figure}[t]
\begin{center}
\includegraphics[width=0.5\textwidth]{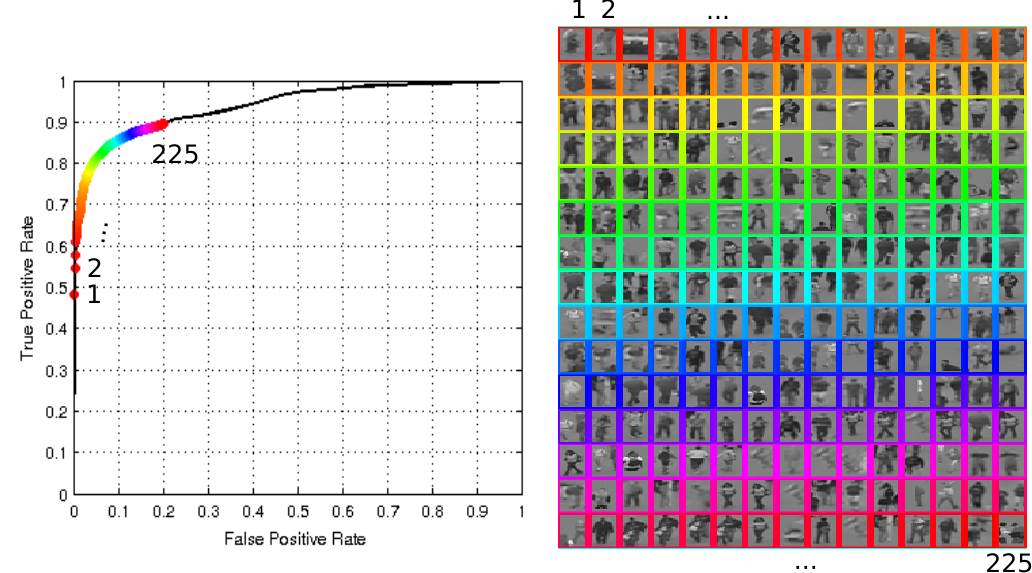}
\end{center}
\caption{Analysis of the false positive instances generated by our video 
parsing on the UCSD \textit{ped1} dataset. Instances are sorted in the 
decreasing order of their abnormality score. \label{fig:ped1false}}
\end{figure}

Our study shows that video parsing outperforms all other methods in  
experiments on both UCSD \textit{ped1} and \textit{ped2} datasets. Fig. 
\ref{fig:ped1roc} shows ROC curves for the frame-wise labeling of the UCSD 
\textit{ped1} set. Tab. \ref{tab:ped1tab} gives the performance measures 
for the \textit{ped1} dataset. We see that the inclusion of the temporal 
component and the improved inference enables spatio-temporal video parsing to 
improve upon our previous sequential video parsing by $2.9\%$ in AUC and  
$5.1\%$ in EER. From Tab. \ref{tab:ped1tab} we also see that our approach 
improves upon recently proposed powerful methods such as LSA \cite{Saligrama12} 
($1.2\%$ gain in AUC and $3.1\%$ in EER) as well as SCL \cite{Lu13} ($2.1\%$ 
gain in AUC and EER). All ROC plots for the pixel-wise labeling on \textit{ped1} 
are shown in Fig. \ref{fig:ped2roc} b) and c). For the partial pixel-wise 
labeling of \textit{ped1}, the spatio-temporal video parsing achieves an 
improvement of $4.7\%$ AUC and $7.2\%$ RD over the sequential video parsing. We 
outperform the closest competitor (HDMT CRT \cite{Li13}) by $14.1\%$ in AUC and 
$10.4\%$ in RD. For the full pixel-wise labeling of \textit{ped1}, we achieve an 
improvement of $2.5\%$ in RD over the sequential video parsing. The competing 
HMDT CRF \cite{Li13} method we outperform in this case by $1.5\%$ in AUC and 
$5.0\%$ in RD.

The ROC curves for the frame-wise labeling of UCSD \textit{ped2} are given in 
Fig. \ref{fig:ped2roc} a). The numerical results are given in Tab. 
\ref{tab:ped2tab}. We observe an improvement in performance of 
spatio-temporal parsing over sequential parsing by $2.6\%$ in AUC and $3.4\%$ 
in EER. The best method so far, MHDT CRF \cite{Li13}, we improve upon by 
$6.9\%$ in EER. For the pixel-wise labeling of \textit{ped2} dataset, we 
outperform the competing HMDT CRF method by $8.7\%$ RD (AUC values for HMDT CRF 
are not provided in \cite{Li13}). Overall we see that our spatio-temporal 
reasoning and the convex optimization based inference yield a significant
improvement over the state-of-the-art.
 
Due to temporal grouping of per-frame object candidates (Sect. 
\ref{sec:shortlist}), spatio-temporal video parsing requires significantly 
less hypotheses (only about a hundred for the whole spatio-temporal domain) 
than sequential video parsing \cite{AnticO11}, which needs the same number of 
hypotheses for representing single frames. Since there remain fewer hypotheses 
to process, spatio-temporal video parsing takes less time to execute than 
sequential video parsing. Our non-optimized Matlab implementation on a 
Dual-Core 2.7GHz CPU runs at about $1$ fps, whereas our previous sequential 
video parsing took $5$-$10$ secs per frame. This is on par with recent H-MDT CRF 
\cite{Li13} and Sparse \cite{Saligrama12} methods, with a notable exception of 
extremely fast  SCL method \cite{Lu13}.

\subsubsection{Analysis of False Detections}

To get a full understanding of the detection performance of proposed video 
parsing, we analyze the false detections on the UCSD \textit{ped1} dataset. In 
Fig. \ref{fig:ped1false} we see the first $225$ false detections sorted in the 
decreasing order of their probability of abnormality. We observe several reasons 
for false detections: i) In many cases, false detections appear as a result of 
artifacts in the foreground segmentation. In such cases, wrongly segmented 
pixels cannot be explained by the learned shape model and thus they are 
classified as abnormal. ii) Large variability of the normal human gait can  
sometimes be interpreted in video parsing as abnormal (e.g. running vs. fast 
walking). iii) Seldom errors in the provided video annotation cause that 
correctly detected abnormalities are sometimes considered as false (e.g. cars 
or running persons in Fig. \ref{fig:ped1false}). iv) When the true-positive 
hypothesis is missing from the shortlist due to a non-maximal recall, video 
parser can select an incorrect hypothesis as a next best fit.

\subsection{Evaluation on the UMN Anomaly dataset}

We additionally evaluate our video parsing on the UMN dataset that is 
widely used for benchmarking abnormality detection. The UMN dataset 
consists of three scenes in which periods of normal activity are followed by 
periods of emergency that are staged by people in the scene. In normal 
cases people are walking around alone or in groups. However, in emergency 
cases people start in panic to run away. For each scene several normal and 
abnormal events are happening one after another. In scene one, two and 
three there are two, six and three abnormal events, respectively. The dataset 
does not provide pixel-wise ground-truth abnormality maps, so we follow the 
standard protocol for this dataset and evaluate the detection results only in a 
frame-wise manner. Fig. \ref{fig:umnframe} a) shows ROC curves for the 
frame-wise labeling. The performance measures AUC and EER are given in Tab. 
\ref{tab:umn_quant}. For scene one, our performance is on par with the best 
competing methods in terms of AUC ($99.5\%$) and EER($3.2\%$). For scene two we 
achieve $97.5\%$ AUC that is equal to the best performing method (Sparse 
\cite{Cong11}). For the scene three we achieve $99.9\%$ AUC that improves upon 
the best competitor (Sparse \cite{Cong11}) by $3.5\%$. A qualitative comparison 
of our method to HMDT CRF \cite{Li13} on two frames is shown in Fig. 
\ref{fig:umnframe} b). We see that our method achieves best localization of 
abnormalities that is consistent with findings from earlier experiments on UCSD 
\textit{ped1} and \textit{ped2}.

\section{Conclusion}
\label{sec:conclusion}

In this paper we have framed abnormality detection as spatio-temporal video 
parsing to circumvent the ill-posed problem of directly searching for 
individual abnormal local image regions. We detect abnormalities by searching 
for a set of spatio-temporal object hypotheses that jointly explain the video 
foreground and which are themselves explained by normal training samples. In 
video parsing we do not independently detect individual hypotheses, but their 
joint layout that collectively describes the objects in the scene. We use MAP
inference in a graphical model to effectively localize abnormalities in video 
and solve it as a convex optimization problem. We have evaluated our approach 
on several challenging datasets, which show that video parsing advances the 
state-of-the-art both in terms of abnormality classification and localization.


%

\appendices

\section{Proof of Lemma \ref{thm:hyp_explanation}}
\label{app:derivation_Jh}

\begin{proof}
The hypotheses explanation $J_\Hyp (\{\Obj_\Hyp, \Mod_\Hyp\}_\Hyp)$ (Eq. 
\ref{eq:J}) can be written as follows,
\begin{alignat}{2} \nonumber
 &J_\Hyp(\{\Obj_\Hyp,\Mod_\Hyp\}_\Hyp) = \sum_\Hyp \Bigl\{ - (1 - \Obj_\Hyp)
\log P(\Obj_\Hyp=0 | \Dsc_\Hyp) \\ \nonumber
&- \Obj_\Hyp \log P(\Obj_\Hyp=1 | \Dsc_\Hyp) + \Obj_\Hyp \cdot \log
Z(\Dsc_\Hyp) \\ \nonumber &+\sum_{\Prot=1}^M \underbrace{\Obj_\Hyp
\cdot \mathbf{1} [ \Mod_\Hyp = \Prot ] }_{=\Lat_{\Hyp,\Prot}} \cdot \Bigl(   
\beta \Delta (\Dsc_\Hyp, \Dsc_\Prot)
\\ \nonumber
&- \log P_\Prot^{loc}( \Loc_\Hyp^\T ) - \log P_\Prot^{vel} \bigl(
\Loc_\Hyp^\T - \Loc_\Hyp^{\T - 1} \bigr) \Bigr) \Bigr\}. 
\end{alignat}

By replacing $\Obj_\Hyp$ with the sum from Eq. \ref{eq:o_h}, we see that 
the hypotheses explanation term $J_\Hyp(\{\Obj_\Hyp, \Mod_\Hyp\}_\Hyp)$  can be 
expressed as a linear function of the parsing indicator $\mathbf{\Lat}$,
\begin{equation}  \nonumber
 J_\Hyp(\mathbf{\Lat}) = \mathbf{\Hpar}^\top \mathbf{\Lat} + \Hpar_0,
\end{equation}
where the parameter vector $\mathbf{\Hpar} = \{ 
\Hpar_{\Hyp,\Prot}\}_{\Hyp,\Prot}$ and scalar $\Hpar_0$ are defined in the 
following way,
\begin{alignat}{4} \nonumber 
  \Hpar_{\Hyp,\Prot} &= - \log P(\Obj_\Hyp=1 | \Dsc_\Hyp ) + \log P(\Obj_\Hyp=0 
| \Dsc_\Hyp ) \\ \nonumber
  &+ \log Z(\Dsc_\Hyp) + \beta \Delta (\Dsc_\Hyp, \Dsc_\Prot) - \log 
P_\Prot^{loc} ( \Loc_\Hyp ) \\ \nonumber
  &- \log P_\Prot^{vel} \bigl( \Loc_\Hyp^\T - \Loc_\Hyp^{\T - 1} ) \\ \nonumber
   \Hpar_0 &= - \sum_\Hyp \log P(\Obj_\Hyp=0 | \Dsc_\Hyp),
\end{alignat}
and they do not depend on the parsing indicator $\mathbf{\Lat}$. 
\end{proof}

\section{Proof of Lemma \ref{thm:convexity}}
\label{app:convexity}
\begin{proof}
The second derivative of the function $\Aux_{\Fg_\Pix^\T} (x), \ x > 0$ is 
given as follows,  
\begin{alignat}{2} \nonumber
  \Aux_{\Fg_\Pix^\T}^ {\prime\prime}(x) &=  \Fg_\Pix^\T \cdot \frac{e^{-x}} {( 1
- e^{-x} )^2}.
\end{alignat}
We see that the second derivative is positive, $\Aux_{\Fg_\Pix^\T} 
^ {\prime \prime}(x) > 0$, if the parameter $\Fg_\Pix^\T$ is positive, 
$\Fg_\Pix^\T > 0$, so in this case the function $\Aux_{\Fg_\Pix^\T}(x)$ is 
strictly convex. If the parameter $\Fg_\Pix^\T$ equals zero, $\Fg_\Pix^\T = 
0$, the function $\Aux_{\Fg_\Pix^\T} (x) $ is linear, $\Aux_{\Fg_\Pix^\T}(x) = 
x$, and therefore convex as well. 
\end{proof}

\section{Proof of Lemma \ref{thm:pix_explanation}}
\label{app:derivation_Jpix}

\begin{proof}
The foreground explanation $J_\Pix (\{\Obj_\Hyp,\Mod_\Hyp\}_\Hyp)$ 
depends on all hypotheses that cover pixel $\Pix$,
\begin{alignat}{2} \nonumber
 J_\Pix &(\{\Obj_\Hyp, \Mod_\Hyp\}_\Hyp) = \\ \nonumber
 &\sum_\Pix \Bigl\{ -(1-\Fg_\Pix^\T) \log P(\Fg_\Pix^\T=0 | \{ \Obj_\Hyp,
\Mod_\Hyp, \Loc_\Hyp\}_\Hyp) \\ \nonumber
 &- \Fg_\Pix^\T \cdot \log \bigl( 1 - P(\Fg_\Pix^\T=0 | \{ \Obj_\Hyp, \Mod_\Hyp,
\Loc_\Hyp\}_\Hyp) \bigr) \Bigr\} \\ \nonumber
 &= \sum_\Pix \Aux_{\Fg_\Pix^\T} \bigl(-\log P(\Fg_\Pix^\T=0 | \{ \Obj_\Hyp,
\Mod_\Hyp, \Loc_\Hyp\}_\Hyp)
\bigr). 
\end{alignat}

The argument of the function $\Aux_{\Fg_\Pix^\T}(\cdot)$ in the last equation 
is bilinear in the parsing indicator $\mathbf{\Lat}$ (Eq. \ref{eq:subst}) and 
the joint shape prototype vector $\bm{\Shape}$ (Eq. \ref{eq:shape_vector}),
\begin{alignat}{2} \nonumber
 &-\log P(\Fg_\Pix^\T=0 | \{ \Obj_\Hyp, \Mod_\Hyp, \Loc_\Hyp\}_\Hyp) \\ 
\nonumber
 &= -\log (1 - P_0) - \sum_\Hyp \log \bigl( 1 - P(\Fg_\Pix^\T=1 | \Obj_\Hyp,
\Mod_\Hyp, \Loc_\Hyp) \bigr) \\ \nonumber
 &= - \log (1 - P_0) -\sum_\Hyp \sum_\Prot \underbrace{\Obj_\Hyp \cdot
\mathbf{1} [ \Mod_\Hyp = \Prot ] }_{=\Lat_{\Hyp,\Prot}} \cdot \mathbf{1} [ \Pix 
\in \mathcal{S}_\Hyp^\T ] \\  \nonumber
 &\cdot  \sum_{\Pix^\prime} \mathbf{1} [  \Loc_\Pix^\T = s_\Hyp^\T
\cdot \Loc_{\Pix^\prime}^\T + (x_\Hyp^\T \ y_\Hyp^\T )^\top ] \cdot
\underbrace{\log P_\Prot (\Fg_{\Pix^\prime}^\T = 0)}
_{=:-\bm{\Shape}_{\Prot,\Pix^\prime}}  \\  \nonumber
 &= \bm{\Shape}^\top \mathbf{ \MakeUppercase{\Xpar} }_\Pix \mathbf{\Lat} +
\Xpar_0,
\end{alignat}
where $\mathbf{ \MakeUppercase{\Xpar} }_\Pix$ is a sparse 
matrix with following elements,
\begin{equation} \nonumber
 \mathbf{ \MakeUppercase{\Xpar} }_\Pix(\Prot,\Pix^\prime;\Hyp,\Prot) = 
\mathbf{1} [ \Pix \in \mathcal{S}_\Hyp^\T ] \cdot \mathbf{1} [ \Loc_\Pix^\T =
s_\Hyp^\T \cdot \Loc_{\Pix^\prime}^\T + (x_\Hyp^\T \ y_\Hyp^\T )^\top ],
\end{equation}
and the scalar $\Xpar_0$ has the value $\Xpar_0 = -\log (1 - P_0)$.

Thus, the foreground explanation term $J_\Pix(\{\Obj_\Hyp, \Mod_\Hyp\}_\Hyp)$ 
can be written as
\begin{alignat}{2} \nonumber
 J_\Pix (\mathbf{\Lat}, \bm{\Shape}) &= \sum_\Pix \Aux_{\Fg_\Pix^\T}
\Bigl( \bm{\Shape}^\top  \mathbf{ \MakeUppercase{\Xpar} }_\Pix  \mathbf{\Lat} +
\Xpar_0 \Bigr).
\end{alignat}
\end{proof}

\section{Proof of Lemma \ref{thm:lipschitz}} \label{app:lipschitz}

\begin{proof}
The expression for the first derivative of the function $\Aux_{\Fg_\Pix^\T} 
(x)$ is 
\begin{alignat}{2} \nonumber
  \Aux_{\Fg_\Pix^\T}^\prime(x) &= 1 - \Fg_\Pix^\T \cdot \frac{1} {1 - e^{-x}},
\end{alignat}
The absolute difference of the first derivative of function 
$\Aux_{\Fg_\Pix^\T} (x)$ evaluated in points $x_1, x_2 \geq \Xpar_0 = -\log 
(1 - P_0)$ is upper bounded in the following way,
\begin{alignat}{2} \nonumber
 &\bigl| \Aux_{\Fg_\Pix^\T}^\prime (x_1) - \Aux_{\Fg_\Pix^\T}^\prime (x_2) 
\bigr| =  \Fg_\Pix^\T \cdot \bigl | \frac{1}{1 - e^{-x_1}}  - \frac{1}{1 - 
e^{-x_2}} \bigr | \\ \nonumber
 &= \Fg_\Pix^\T \frac{ \bigl| e^{-x_1} - e^{-x_2} \bigr|}{(1 - e^{-x_1}) (1 -
e^{-x_2})} \leq \Fg_\Pix^\T \frac{1}{P_0^2} \bigl| e^{-x_1} - e^{-x_2} \bigr| \\
\nonumber
 &= \Fg_\Pix^\T \frac{1}{P_0^2} \cdot e^{-\min\{ x_1, x_2 \}} \cdot \bigl( 1 -
e^{-|x_1 - x_2|} \bigr) \\  \nonumber
 &\leq \Fg_\Pix^\T \frac{1-P_0}{P_0^2} \cdot |x_1 - x_2|  = \rho |x_1 - x_2|.
\end{alignat}
In the last line of the proof we used the inequality \\ $1 - e^{-x} \leq x, \ 
\forall x > 0$. 
\end{proof}

%

\ifCLASSOPTIONcaptionsoff
  \newpage
\fi



\bibliographystyle{IEEEtran}
\bibliography{egbib.bib}
\end{document}